# Color wave: A systematic evaluation of active screen illumination for injection attack detection in remote face authentication

Daniyar Kurmankhojayev[1], Elmurat Nigmatov, Dmitrii Gordin, Andrei Shadrikov, Mikhail Shkorin, Danijar Gabdullin, Vyacheslav Shen

Department of Research and Development, Verigram LLP, Almaty, Kazakhstan

**Abstract**

Remote face authentication requires protection against both presentation attacks and biometric data injection attacks prior to face matching. Methods based on active screen illumination appear promising for both by introducing randomized visual stimuli through the user's device screen and evaluating the corresponding facial responses. Although such methods have been more studied for presentation attack detection (PAD), their systematic evaluation for injection attack detection (IAD) remains limited. To address this gap, we first classify injected content into static images, prerecorded videos, and real-time deepfakes based on their expected responses to the challenge. We then adapt an active screen-illumination approach into an implementation we term *color wave* and develop a systematic, injection-specific evaluation protocol. The protocol examines detection effectiveness, generalization across operating conditions, robustness to signal degradation, and susceptibility to selected evasion strategies. The findings demonstrate a strong discriminative signal under controlled conditions but also reveal performance limitations under overexposure, low screen brightness, and subject motion, as well as vulnerabilities to selected evasion strategies. By relating detection outcomes to attack types and operating conditions, this study provides an interpretable assessment of when and why active screen illumination succeeds or fails as an IAD mechanism.

[1] Corresponding author: d.kurmankhojayev@verigram.kz

# 1. Introduction

## 1.1 Background and motivation

Remote face authentication must verify not only the identity of a subject but also the authenticity of the acquired biometric sample, as the acquisition process is vulnerable to two distinct threats: presentation attacks and biometric data injection attacks *(Yu et al., 2023; Carta et al., 2024)*. In presentation attacks, an adversary presents an artifact or manipulated biometric characteristic to a physical camera *(Ramachandra & Busch, 2017)*. In biometric data injection attacks, the adversary replaces or modifies the video stream received by the authentication system, potentially bypassing the camera entirely *(Carta et al., 2022)*. By bypassing the physical camera, biometric data injection attacks may evade conventional presentation attack detection (PAD), motivating complementary injection attack detection (IAD) mechanisms *(Carta et al., 2023)*.

Methods based on active screen illumination offer a potential solution by introducing randomized, session-specific visual stimuli and examining the corresponding facial responses *(Smith et al., 2015; Gerstner & Farid, 2022)*. For PAD, both bona fide faces and physical attack instruments, such as printed photographs and displays, may respond to illumination changes. Therefore, color correspondence alone may be insufficient, requiring additional evidence from structured illumination patterns, shadows, and reflections to distinguish three-dimensional facial geometry from flat attack instruments *(Chan et al., 2018; Tang et al., 2018)*. For IAD, the video stream bypasses physical camera acquisition, making correspondence between the observed facial color response and the current, randomized illumination stimulus potentially sufficient to distinguish bona fide input from certain injection attacks, without requiring three-dimensional facial cues. This may simplify method development by eliminating the need to design structured illumination patterns and reducing reliance on extensive representative datasets for training models to capture spatial facial responses.

However, the effectiveness of this approach for IAD cannot be inferred directly from PAD studies. While static images and prerecorded videos generally lack responses corresponding to the current stimulus, real-time deepfakes may reproduce these responses, albeit with substantial processing delays, making temporal correspondence an important additional detection cue *(Gerstner & Farid, 2022)*. Although previous studies have investigated active illumination against real-time face manipulations, they have largely focused on video conferencing *(Shang & Wu, 2020; Gerstner & Farid, 2022)*, where unobtrusive verification and uninterrupted interaction are often prioritized *(Liu et al., 2020; Shang & Wu, 2020)*. Similarly, their findings cannot be directly generalized to remote face authentication, which can accommodate more conspicuous challenges and where security failures may have more severe consequences, including unauthorized access and identity fraud. Furthermore, adaptive evasion strategies may detect the current stimulus color and modify the injected video accordingly, potentially passing color-based checks. These differences and vulnerabilities motivate a systematic, authentication-oriented evaluation of detection effectiveness across attack scenarios and operating conditions.

## 1.2 Literature review

Prior work distinguishes injection attacks by delivery method and injected-content characteristics. This distinction is formalized in *CEN/TS 18099:2024*, which characterizes biometric data injection attacks through the combination of an Injection Attack Method (IAM) and an Injection Attack Instrument (IAI). In remote face authentication, injection methods include virtual-camera substitution and direct injection into driver or application-level pipelines *(Carta et al., 2022, 2023, 2024)*. The standard distinguishes unmodified, modified, and artificial biometric samples. In this study, we introduce a separate classification of injected content into static images, prerecorded videos, and live content (including real-time deepfakes), based on their expected responses to a session-specific illumination challenge. The same content can be delivered through different injection methods.

IAD approaches can be broadly classified as *metadata-based* or *content-based*. Metadata-based methods identify potential injection by examining camera-interface or session characteristics rather than visual content. *Kurmankhojayev et al. (2025)* detect virtual cameras through camera-interface behavior, while *Carta et al. (2024)* review direct injection techniques and associated countermeasures. Content-based methods instead analyze the received video and are, in principle, less dependent on the injection channel. They may be *passive* or *active*, using visual, photometric, behavioral, or physiological cues. Passive methods may rely on manipulation-specific artifacts, potentially limiting generalization *(Khan et al., 2025; Lai et al., 2026)*. Related studies highlight challenges in detecting facial manipulations under low-resolution conditions and generalizing across deepfake datasets *(Agarwal & Ratha, 2025; Lai et al., 2025)*. Active methods instead introduce unpredictable, session-specific challenges, whose effectiveness depends on challenge design and attacker capabilities *(Gerstner & Farid, 2022; Mittal et al., 2024; Firc et al., 2026)*. Among various challenge–response methods, *active screen illumination* is particularly promising because, unlike movement-based challenges, it elicits an almost immediate physical response without requiring deliberate user action, while reproducing this response in injected video may introduce detectable delays *(Tang et al., 2018; Gerstner & Farid, 2022)*.

Active illumination has been investigated for PAD, where the emphasis is on distinguishing genuine three-dimensional faces from physical attack instruments. *Chan et al. (2018)* examine flash-based facial reflectance to detect two-dimensional presentation attacks, while *Tang et al. (2018)* employ randomized illumination patterns to exploit facial geometry, reflections, and response timing. Related challenge-based authentication methods use camera movement or head rotation to capture spatial and temporal facial characteristics *(Castelblanco et al., 2022)*. In the context of IAD for remote face authentication, *Smith et al. (2015)* investigate virtual-camera replay attacks and propose a countermeasure based on colored facial reflections. More recently, *Vijaykumar et al. (2024)* examine screen illumination for detecting deepfake injection attacks in biometric recognition. Related studies in video conferencing investigate illumination-based detection of replayed and manipulated video. *Liu et al. (2020)* propose unobtrusive illumination for detecting replay attacks, *Shang and Wu (2020)* employ brightness modulation against face reenactment, and *Gerstner and Farid (2022)* investigate colored illumination against real-time deepfakes. *Guo et al. (2023)* instead use structured screen patterns and corneal reflections for active probing. Broader challenge–response approaches include interactive challenges against real-time deepfakes *(Mittal et al., 2024)* and randomized facial gestures against live face swaps *(Firc et al., 2026)*. These studies investigate different attack scenarios and detection cues, with some video-conferencing methods emphasizing minimal interference with normal interaction. Commercial systems employ active illumination to address both PAD and IAD. Amazon Rekognition Face Liveness uses colored screen illumination and claims protection against presentation and digital injection attacks *(Amazon Web Services, 2023)*. Similarly, iProov's Flashmark technology employs randomized color sequences to verify facial responses, with applications to both presentation and injection attack detection *(iProov, n.d.; Bud, 2018; Bud & Newell, 2025)*.

Although previous research demonstrates the potential of active illumination, differences in attack coverage, application context, and evaluation methodology limit direct comparisons of its findings. Despite the general IAD evaluation guidance provided by *CEN/TS 18099:2024*, systematic evaluations of active screen illumination for remote face authentication remain limited, particularly regarding generalization, signal degradation, and adaptive evasion. A unified evaluation protocol is therefore needed to characterize detection effectiveness across attack scenarios and operating conditions while identifying the factors underlying detection successes and failures.

## 1.3 Objectives and contributions

This study develops a systematic, injection-specific evaluation protocol for active screen illumination in remote face authentication. It focuses on content-based IAD and adapts the active screen-illumination principle into an implementation termed color wave. Because the method analyzes received video rather than injection-channel metadata, we assume that its performance should not substantially depend on whether the stream is delivered through a virtual camera or direct application-level injection. The experiments presented here evaluate the virtual-camera case.

The study pursues three objectives and corresponding methodological and empirical contributions:

1. **Threat characterization, implementation, and evaluation protocol.** Classify injected content into static images, prerecorded videos, and real-time deepfakes based on their expected responses to illumination stimuli. Define an authentication-oriented threat model specifying injection channels, attacker capabilities, and security assumptions. Develop a reproducible color wave implementation and a structured evaluation protocol with explicit challenge generation, response analysis, decision criteria, and experimental conditions.

2. **Detection performance and generalization.** Evaluate the ability of stimulus–response color agreement, duration matching, and temporal delay to distinguish bona fide sessions from static images, prerecorded videos, and real-time deepfakes. Assess generalization across users, devices, environments, illumination, motion, virtual-camera software, and deepfake implementations.

3. **Operating limits, interpretability, and evasion.** Systematically characterize performance under signal degradation, including adverse illumination and stream-quality conditions, and susceptibility to selected evasion strategies. Relate detection successes and failures to variations in attack types, operating conditions, and observed facial-response characteristics, identifying the factors underlying performance degradation and potential vulnerabilities to adaptive attacks.

Together, these contributions provide a systematic and interpretable empirical assessment of active screen illumination for IAD, identifying the conditions under which detection succeeds or fails and the limitations of the approach.

The remainder of the manuscript is organized as follows. Section 2 describes the threat model, color wave method, decision criteria, evaluation protocol, and data collection. Section 3 presents the experimental evaluation, covering method configuration, baseline validation, generalization, signal degradation, evasion strategies, and ablation analysis, followed by a discussion of the findings and limitations. Section 4 concludes the paper and outlines directions for future research. The Appendix details the calibration experiments used to select the stimulus colors, durations, and response-color predictor.

# 2. Methodology

This section presents the methodology of the study. Section 2.1 characterizes the considered injection attack vectors and injected-content classes. Section 2.2 describes the developed color wave method, from challenge generation and facial-response analysis to session classification. Section 2.3 presents the evaluation protocol and corresponding data-collection procedure.

## 2.1 Injection attack taxonomy

This study adopts the Injection Attack Method (IAM)–Injection Attack Instrument (IAI) framework established in *CEN/TS 18099:2024*, while introducing a different grouping of IAIs tailored to active screen-illumination challenges. Whereas the standard classifies IAIs as unmodified, modified, or artificial biometric samples, we distinguish three temporal classes—static images, prerecorded videos, and live injected streams—based on their expected responses to the *color wave* stimulus. This complementary classification enables attacks with similar stimulus–response characteristics to be evaluated together, regardless of how the injected content is generated.

Fig. 1 illustrates the distinction between injection methods and injected-content classes. The upper layer identifies two IAMs: virtual-camera injection, where software or drivers substitute the physical camera feed, and direct injection, where biometric data are replaced or modified within the acquisition or application pipeline. The lower layer presents the three temporal IAI classes and their corresponding attack types. Either IAM can deliver content from any of these classes.

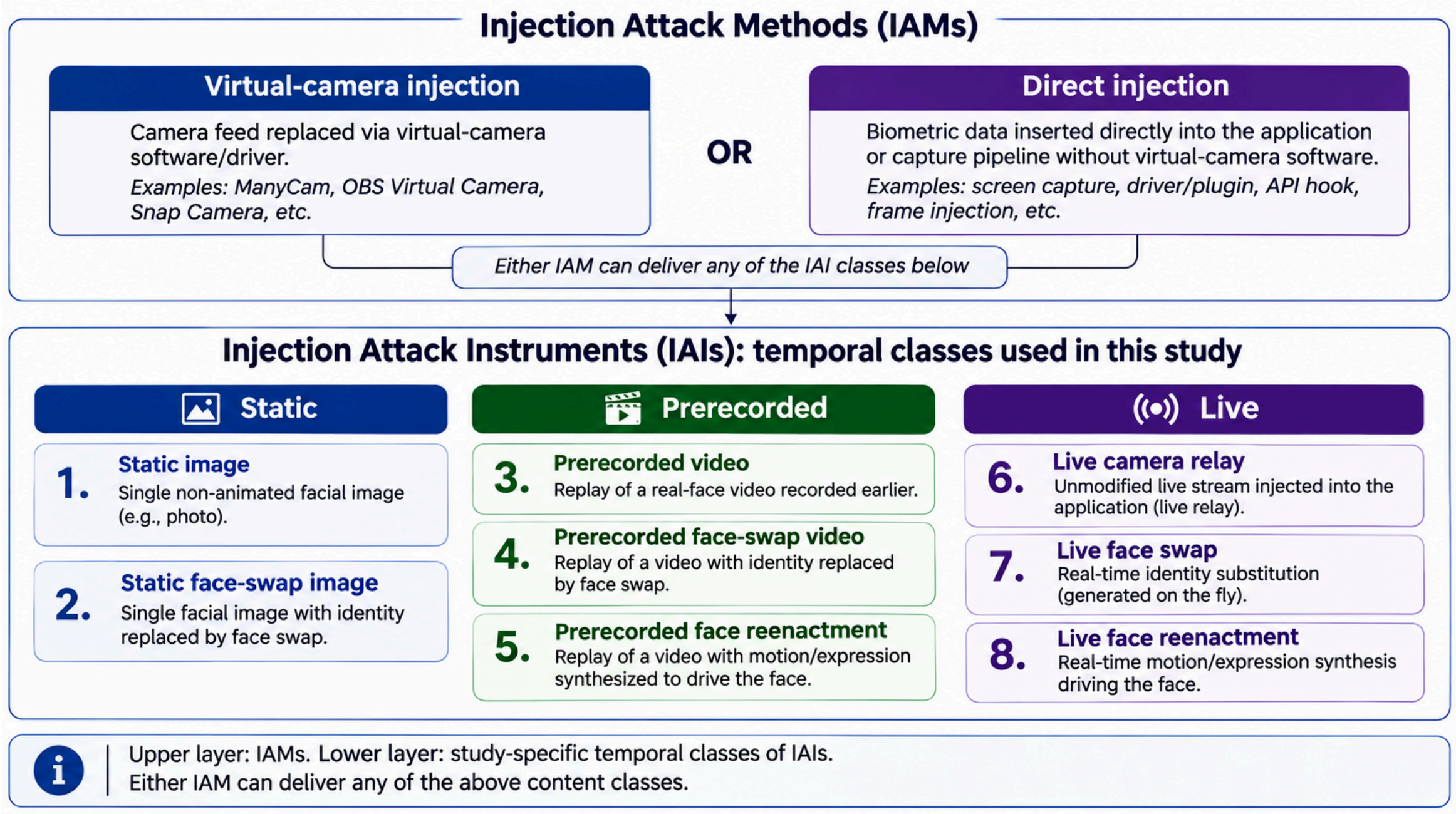

**Figure 1.** Taxonomy of biometric data injection attacks by injection channel and injected-content class.

The three classes differ in their temporal characteristics and potential responses to the randomized illumination stimulus. Static images exhibit no temporal variation, whereas prerecorded videos may contain facial color changes that are not synchronized with the current, session-specific stimulus. Although distinguished in the taxonomy, static-image attacks are treated as a special case of prerecorded-video attacks in the evaluation, where the injected stream consists of identical frames. Both are therefore evaluated together because they generally lack responses corresponding to the current challenge. In contrast, live injected streams may preserve or reproduce stimulus-dependent facial responses, although real-time manipulation and processing may introduce delays or photometric inconsistencies. This distinction allows the evaluation to focus on differences in stimulus–response behavior while retaining a more detailed characterization of the injected content. Although the broader threat model encompasses both IAMs, the experimental evaluation in this study focuses on attacks delivered through virtual-camera software.

## 2.2 Color wave method

Fig. 2 illustrates the overall workflow of the color wave method for injection attack detection, formulated as a binary classification problem. The method relies on the stimulus–response concept (Section 2.2.1). First, the device presents a randomized sequence of color stimuli with varying durations (Section 2.2.2), while the camera captures and synchronizes the facial video (Section 2.2.3). Next, the recorded responses are analyzed to estimate temporal delays (Section 2.2.4) and predict facial-response colors (Section 2.2.5). The predicted response sequence is then compared with the known stimulus sequence to evaluate color agreement (Section 2.2.6) and response-duration matching (Section 2.2.7). Finally, the session is classified as bona fide only if all three criteria—color agreement, duration matching, and session-specific delay—are satisfied; otherwise, it is classified as an attack (Section 2.2.8).

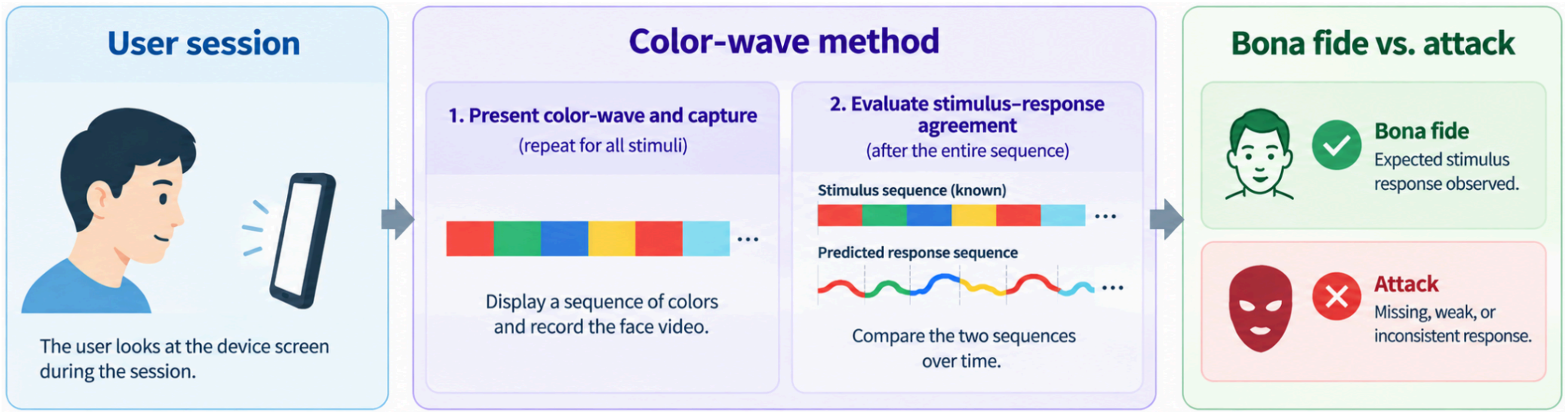

**Figure 2.** Overview of the color wave method for injection attack detection.

## 2.2.1 Stimulus-response concept

Fig. 3 illustrates the expected facial response to a single color stimulus during bona fide acquisition. When the device screen displays a color, the emitted light illuminates the face, producing a measurable change in facial color. However, the observed response may begin after a short delay, develop gradually, and persist beyond the stimulus duration before returning to its baseline. Consequently, the onset and duration of the measured response may differ from those of the stimulus.

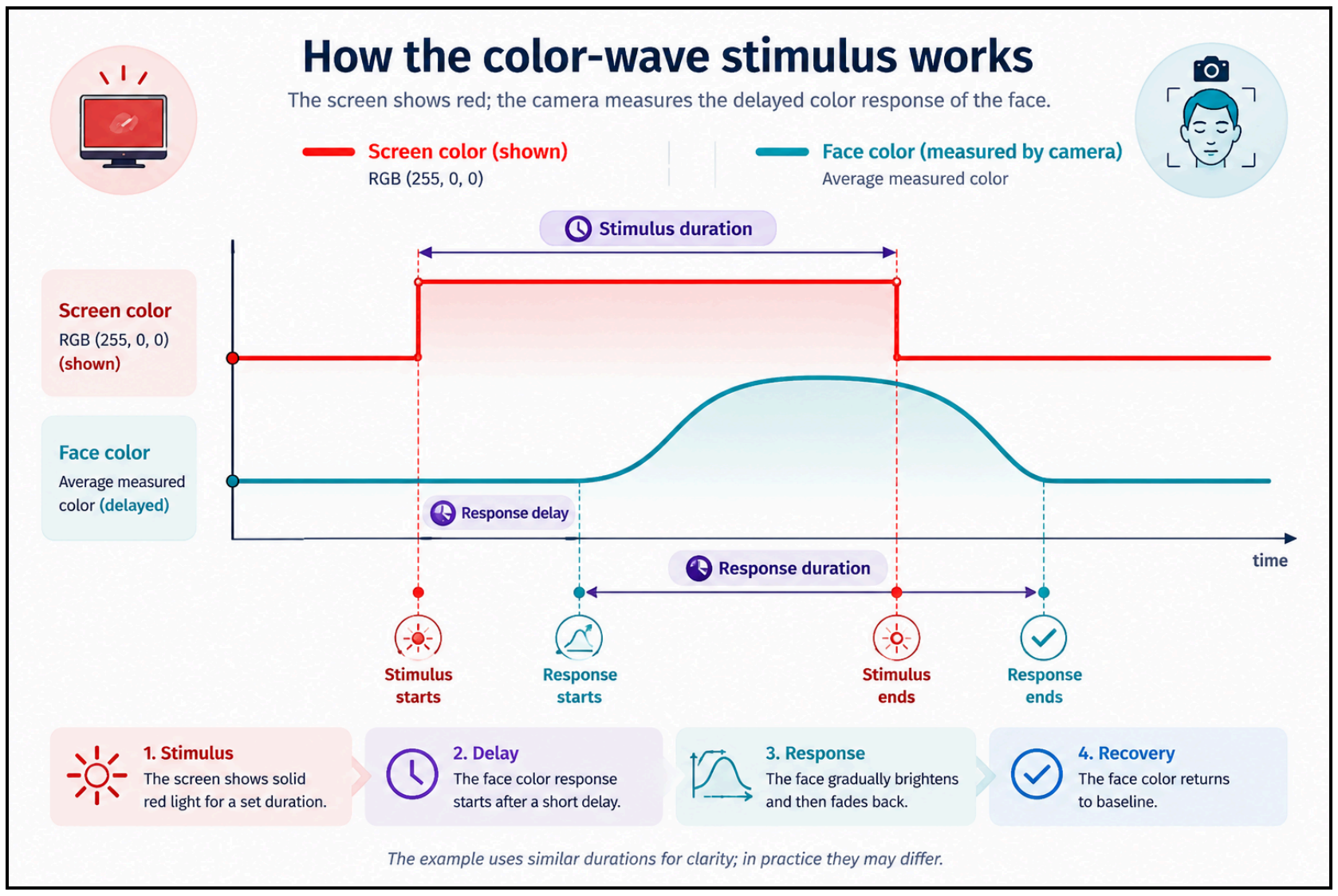

**Figure 3.** Conceptual stimulus–response behavior for a single color stimulus.

## 2.2.2 Active illumination challenge design

The challenge design addresses two major threats: replay attacks (including static image attacks as a special case) using recordings of previously successful bona fide sessions and live injection attacks involving real-time deepfakes. Previously recorded sessions contain facial responses to earlier illumination challenges and could pass verification if the same challenge were reused. Real-time deepfakes, in contrast, may reproduce facial responses to the current stimulus, but the required processing may introduce delays or inconsistencies. To address these threats, the *color wave* method generates a randomized, session-specific challenge, making previously recorded responses unlikely to match the current challenge while providing an unpredictable stimulus for evaluating live manipulations.

The challenge is defined by four parameters:

1. *Color palette:* The predefined colors available for generating the challenge.
2. *Number of stimuli:* The number of consecutive color stimuli presented during a session.
3. *Operational stimulus-duration set:* The predefined set of durations that can be assigned to individual stimuli.
4. *Ordering rule:* The procedure for selecting and arranging stimulus colors and durations.

For each session, colors are randomly selected from the color palette without replacement, while durations are assigned from the operational stimulus-duration set $\mathcal{D}$, allowing repetition but ensuring that each available duration appears at least once. This randomization increases challenge diversity and reduces the likelihood that previously recorded facial responses match the current challenge. However, it does not guarantee resistance to adaptive attacks capable of reproducing the expected responses.

Let $C$ denote the size of the color palette, $D$ the size of the discrete stimulus-duration choice set, and $M$ the number of stimuli. The number of possible challenges is:

$$K = \frac{C!}{(C-M)!} \sum_{k=0}^{D} (-1)^k \binom{D}{k} (D-k)^M \tag{1}$$

where $M \leq C$ and $M \geq D$. The first term counts the possible ordered color selections, while the second counts duration sequences in which every available duration appears at least once.

The design parameters balance challenge diversity, response distinguishability, reliable acquisition, and session duration. The color palette should contain colors that produce distinguishable facial responses, while the discrete stimulus-duration choice set should ensure sufficient response capture without unnecessarily prolonging the session. The calibration experiments used to select the color palette and stimulus-duration choice set are presented in Appendices A1 and A2, respectively.

### 2.2.3 Frame capture and synchronization

While the stimulus sequence is presented, the camera continuously captures frames. Each frame is processed through face detection, normalization, and cropping to obtain a consistent facial region for subsequent response analysis. Simultaneously, the system records stimulus and frame timestamps using a common time reference. For each stimulus $i$, its start time $t_i^{start}$ and end time $t_i^{end}$ are recorded, while each captured frame $j$ is assigned a timestamp $t_j^{frame}$.

$$\Delta t_{ij} = t_j^{\mathrm{frame}} - t_i^{\mathrm{start}} \quad (2)$$

This synchronization allows each processed facial frame to be positioned relative to the stimulus sequence, providing the temporal information required for subsequent color and timing analysis.

### 2.2.4 Response delay estimation

Response delay is estimated by identifying the temporal shift that maximizes agreement between the displayed illumination stimulus and the observed camera response. For each captured frame, the sRGB image is converted to linear RGB, and the central 60% of the frame is spatially averaged to obtain a three-channel RGB vector. For each candidate delay, an ordinary least-squares regression with an intercept relates the measured camera RGB values to the time-shifted screen RGB values. The delay maximizing the pooled coefficient of determination ($R^2$) across the three color channels is selected. A similar principle of temporal alignment between a known illumination signal and recorded video is employed by *Michael et al. (2025)*, although their approach uses cross-correlation rather than regression-based estimation. Candidate delays range from 0 to 750 ms in increments of 10 ms. The upper bound of 750 ms is a pragmatic choice rather than an empirically established limit.

The method employs two delay quantities estimated using the same procedure but serving different purposes. The reference delay $L_{\mathrm{ref}}$ is estimated exclusively from bona fide calibration sessions and provides a fixed temporal alignment for extracting training frames (Section 2.2.5) and defining the response windows used in color-agreement (Section 2.2.6) and duration-matching evaluations (Section 2.2.7). The session-specific delay $L_s$ is estimated independently for each bona fide or attack session and serves as one of the classification criteria (Section 2.2.8).

The two quantities are kept separate because attack sessions, particularly those involving real-time deepfakes, may exhibit substantially different or inconsistent response delays. Using $L_s$ for response-window alignment could compensate for attack-induced delays and obscure their detection. Therefore, $L_{\mathrm{ref}}$ remains fixed for subsequent response analysis, while $L_s$ is evaluated independently against the delay criterion. If a reliable session-specific estimate cannot be obtained, the session fails this criterion and is classified as an attack.

### 2.2.5 Response color prediction

The response-color predictor aims to identify changes in facial color induced by the illumination stimulus. Since facial appearance varies across subjects, devices, and acquisition conditions, relying on absolute facial color may be unreliable. To account for these variations, the method compares facial observations captured under neutral illumination with those captured during the colored-stimulus sequence. A CNN-based classifier is employed to learn spatially varying color-response features from these observations, rather than relying on predefined color measurements. Related studies have demonstrated the use of learned models for analyzing facial color and reflectance characteristics *(Smith et al., 2015; Shu et al., 2023)*. Fig. 4 illustrates the resulting prediction process.

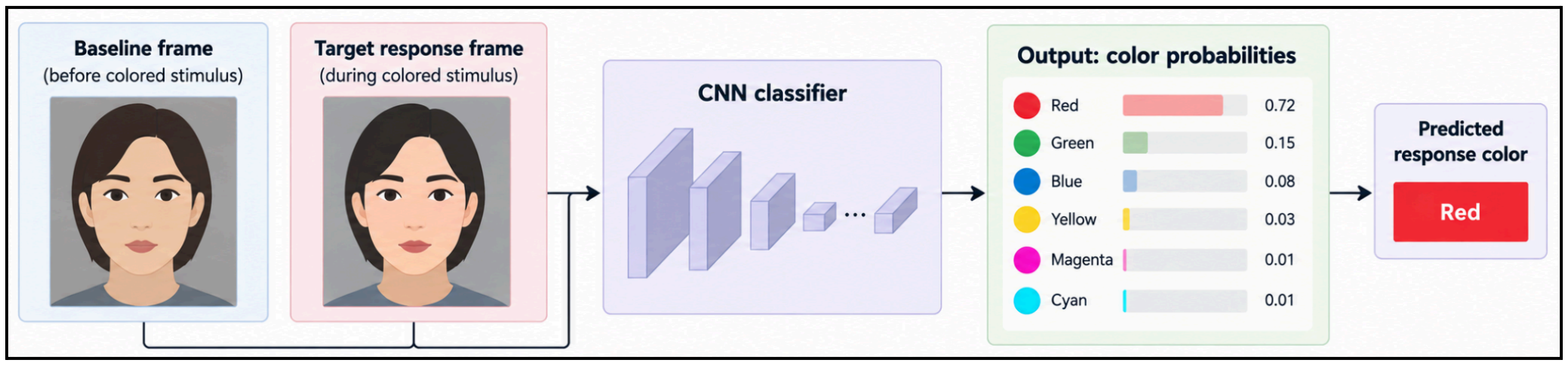

**Figure 4.** Prediction of facial color response using neutral and illuminated facial observations.

For a neutral reference frame $I_b$ and a facial-response frame $I_{r,j}$, the predictor produces a probability vector over the $C$ possible response colors:

$$p_j = f(I_b, I_{r,j}) = [p_{j,1}, p_{j,2}, \dots, p_{j,C}] \quad (3)$$

where $p_{j,c}$ denotes the predicted probability of response color $c$. The class with the highest predicted probability is selected as the response-color label $\hat{c}_j$. Each prediction is associated with its frame timestamp and subsequently used for stimulus–response agreement evaluation (Section 2.2.6) and response-duration matching (Section 2.2.7).

The predictor is trained on bona fide calibration sessions with known stimulus colors and timestamps. Fig. 5 illustrates the frame-selection and labeling procedure.

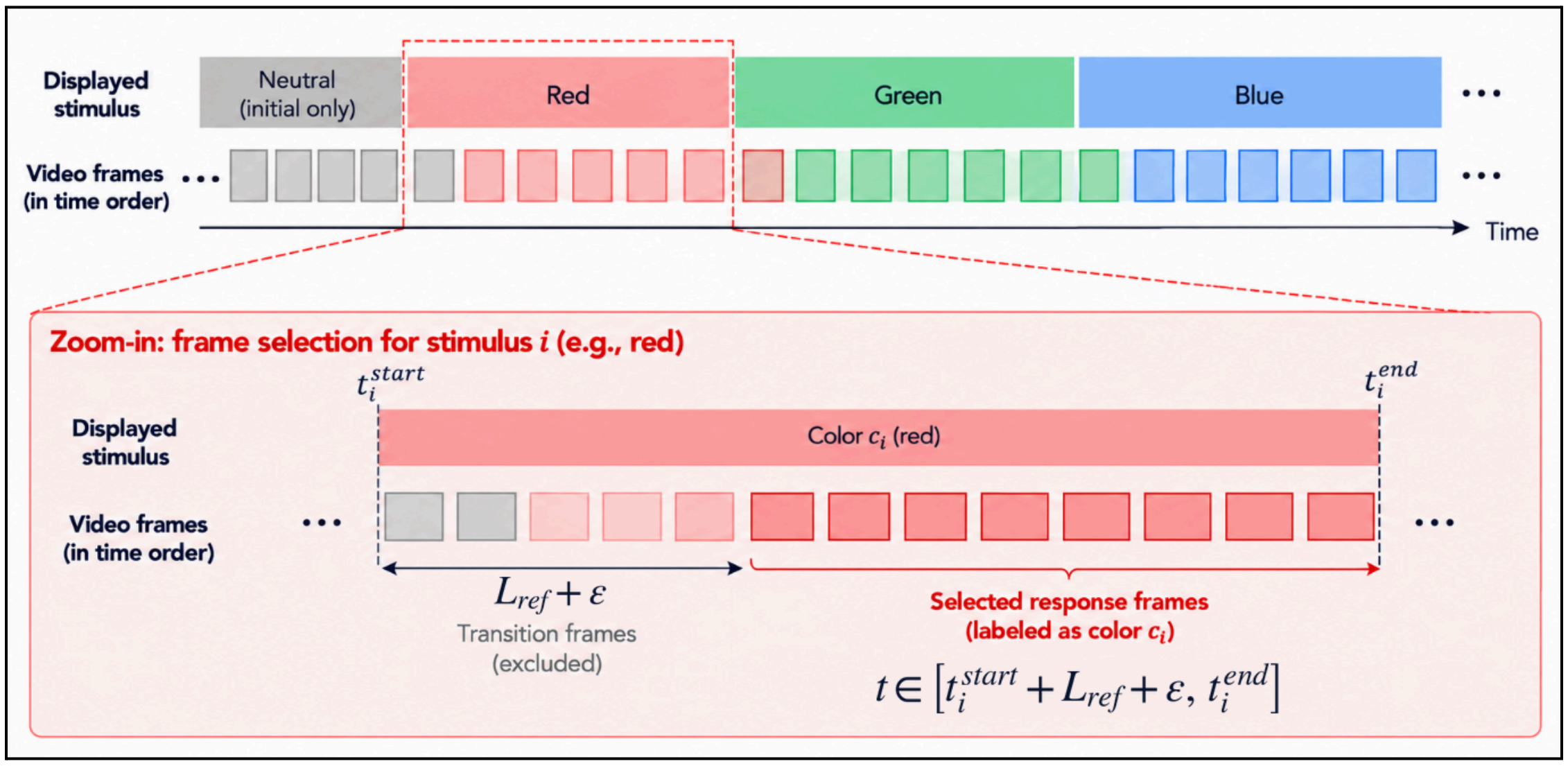

**Figure 5.** Selection and labeling of response frames for training data preparation.

Training frames are selected using the fixed reference delay $L_{ref}$ (Section 2.2.4). For stimulus $i$, with onset time $t_i^{start}$ and end time $t_i^{end}$, response frames are selected from

$$t \in [t_i^{start} + L_{ref} + \varepsilon, t_i^{end}] \quad (4)$$

where $\varepsilon$ is a temporal margin that excludes transitional frames. Selected frames are automatically labeled with the corresponding stimulus color and randomly paired with neutral frames, which are selected using the same principle. Further details of model training and evaluation are provided in Appendix A3.

### 2.2.6 Stimulus-response agreement evaluation

The color wave challenge relies on the correspondence between the displayed stimuli and the resulting facial color response. While bona fide recordings are expected to exhibit this correspondence, injected content may fail to reproduce the session-specific stimulus sequence. Stimulus–response agreement therefore provides a measure of consistency between the observed facial response and the presented challenge.

Stimulus-response agreement is quantified using the probability assigned by the response-color model to the expected stimulus color. Unlike binary agreement based only on discrete color predictions, this probability-based score preserves the model's confidence in the expected response.

Using the delay-estimation procedure described in Section 2.2.4, stimulus-response agreement and response-duration matching use the fixed empirical reference delay $L_{ref}$. For stimulus $i$, presented from $t_i^{start}$ to $t_i^{end}$, the expected response interval is

$$[t_i^{start} + L_{ref}, t_i^{end} + L_{ref}] \quad (5)$$

This interval approximates the expected facial response. Although the actual response onset and offset may deviate slightly from these nominal boundaries, the selected stimulus durations provide sufficient frames representing the developed response, limiting the influence of transitional frames on the agreement score.

Let $c_i$ denote the color of stimulus $i$, and let $V_i$ denote the set of valid frames captured within its expected response interval. A frame is valid when the face is detected and the corresponding response-color prediction is available. The session-level agreement score aggregates the expected color probabilities across all valid frames in the color wave sequence:

$$A_{\mathrm{s}} = \frac{\sum_{i=1}^{M} \sum_{j \in V_i} p_j(c_i)}{\sum_{i=1}^{M} |V_i|} \tag{6}$$

where $M$ is the number of stimuli and $p_{j,c_i}$ is the probability assigned to the expected color $c_i$ for frame $j$.

The resulting score measures overall stimulus-response agreement across the complete color wave sequence, with higher values indicating stronger model support for the expected facial color responses. Differences between the recorded and current stimulus sequences, particularly in replay-based injection attacks, may reduce this agreement and thus provide evidence of injected content.

### 2.2.7 Response duration matching

A bona fide facial response is expected to match the displayed stimulus color for approximately the duration of the corresponding stimulus. Color agreement alone does not fully capture this temporal correspondence, as matching-color frames may occupy only part of the expected response interval. Response-duration matching therefore assesses whether the cumulative duration of matching-color frames is consistent with the nominal stimulus duration.

Direct comparison of matching-frame counts is inappropriate because cameras may operate at different frame rates and the challenge contains stimuli of different durations. To ensure consistency, matching-frame counts are converted to milliseconds using the session's effective frame rate and mapped to the closest value in the common operational duration set.

For each stimulus $i$, the expected response interval is defined by (5). Let $c_i$ denote the displayed color, $V_i$ the valid frame indices within this interval, $\hat{c}_j$ the predicted response-color label for frame $j$, $FPS_s$ the effective frame rate, and $d_i$ the nominal stimulus duration. Duration matching proceeds as follows:

1. *Identify matching frames.* Retain the valid frames whose predicted color matches the displayed stimulus color.

$$\mathcal{F}_i = \{ j \in V_i : \hat{c}_j = c_i \} \tag{7}$$

2. *Estimate the matching-color duration.* Convert the number of matching frames to milliseconds using the session's effective frame rate.

$$\hat{d}_i = \frac{|\mathcal{F}_i|}{FPS_s} \times 1000 \text{ ms} \tag{8}$$

3. *Determine duration agreement.* Map the measured duration to the nearest value in the operational duration set $\mathcal{D}$. A stimulus is duration-matched if the mapped duration equals its nominal duration.

$$\tilde{d}_i = \arg\min_{d \in \mathcal{D}} |\hat{d}_i - d|, \qquad q_i = \mathbf{1}\{\tilde{d}_i = d_i\} \tag{9}$$

where $D$ is the operational duration set.

4. *Aggregate duration matches.* Count the number of duration-matched stimuli across the challenge:

$$D_s = \frac{1}{M} \sum_{i=1}^{M} q_i \tag{10}$$

where $M$ is the total number of stimuli in the color wave sequence.

If any expected response interval contains fewer than five valid frames, the minimum-frame requirement is not satisfied, and the session is classified as an attack. Otherwise, the number of duration-matched stimuli characterizes the temporal consistency of the observed facial response with the displayed challenge. Frame-rate normalization reduces dependence on camera sampling rates, while mapping to the operational duration set accommodates differences between measured and nominal durations. However, capture effects such as dropped frames and motion may still affect duration estimates.

### 2.2.8 Decision rule

The final classification combines three complementary criteria to assess whether the observed facial response is consistent with the presented color wave challenge. For each session $s$, the method evaluates the estimated session-specific response delay $\hat{L}_s$, the session-level stimulus-response agreement score $A_s$ defined in (6), and the number of duration-matched stimuli $D_s$ defined in (10).

A session is classified as bona fide when all three criteria satisfy their respective empirically determined thresholds:

$$\hat{y}_s = \begin{cases} \text{bona fide}, & \tau_L^{min} \geq \hat{L}s \leq \tau_L^{max} \wedge A_s \geq \tau_A \wedge D_s \geq \tau_D, \\ \text{attack}, & \text{otherwise}, \end{cases} \tag{11}$$

where $\tau_L^{min}$ and $\tau_L^{max}$ denote the lower and upper bounds of the acceptable response-delay interval, respectively, while $\tau_A$ and $\tau_D$ are the thresholds for stimulus–response agreement and duration matching.

Thus, failure to satisfy any criterion is sufficient for classifying a session as an attack. Sessions that do not meet the minimum valid-frame requirement are also classified as attacks. All decision thresholds were established using calibration data and fixed before the four evaluation stages described in Section 2.3.1. They remained unchanged throughout these experiments.

## 2.3 Method evaluation protocol and data collection

The proposed evaluation protocol is designed around the method's central hypothesis: a bona fide session produces a facial response that is physically and temporally linked to the current screen challenge, whereas injected content may lack this response, reproduce it inaccurately, or be deliberately adapted to imitate it. Establishing the method's effectiveness therefore requires more than a single controlled comparison. The evaluation examines four progressively more demanding properties: baseline discrimination, generalization, degradation robustness, and evasion resistance. Together, these stages assess both the discriminative capability of the method and its operating limits.

Evaluating these properties requires dedicated data collection because the relevant evidence is session-specific. Each session must pair the presented challenge with the received facial stream and provide a known ground-truth label. Existing benchmark datasets generally lack the synchronized challenge–response data and controlled injection conditions needed for this evaluation. Accordingly, we collect bona fide sessions, source material for constructing attacks, and attack sessions generated under fresh challenges. Under the adopted threat model, replayed or modified biometric data remain attacks even when adapted to the current stimulus; their acceptance is therefore treated as a false acceptance.

The evaluation protocol and corresponding data-collection procedure are described in Sections 2.3.1 and 2.3.2, respectively.

### 2.3.1 Method evaluation protocol

The proposed evaluation protocol comprises four stages, as illustrated in Fig. 6. Method validation establishes whether the stimulus–response signal distinguishes bona fide sessions from the primary attack classes under controlled conditions. Generalization evaluates performance across new subjects, devices, environments, and attack implementations. Degradation robustness examines the effects of illumination variation and reduced stream quality. Finally, evasion resistance assesses attacks deliberately adapted to reproduce or manipulate the expected facial response.

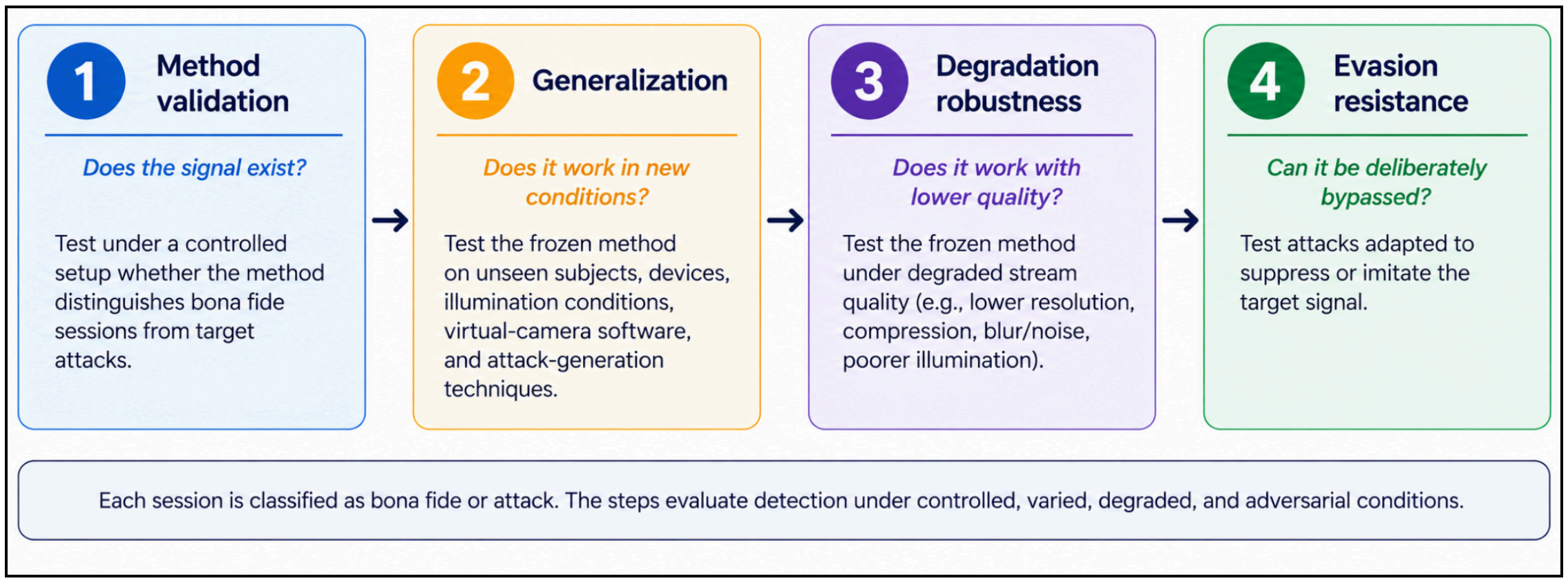

**Figure 6.** Four-stage method evaluation framework.

Table 1 summarizes the experimental factors considered at each evaluation stage.

**Table 1.** Evaluation stages and experimental factors.

| Evaluation stage | Data and experimental factors |
|---|---|
| Method validation | Bona fide sessions compared with static-image, prerecorded-video, and live deepfake attacks under controlled baseline conditions. |
| Generalization (bona fide) | Variation in subjects, capture devices, browsers, backgrounds, subject movement, and lighting conditions. |
| Generalization (attack) | Variation in virtual camera software, media format, deepfake method, and replacement identity. Static images are treated as repeated-frame prerecorded content. |
| Degradation robustness | Variation in screen brightness, ambient-light intensity, frame resolution, and frame rate. |
| Evasion resistance | Prerecorded genuine-response videos evaluated under accelerated replay and manual or automatic stimulus-conditioned frame switching or recoloring. |

The response-color model, challenge-generation configuration, and decision thresholds were established using calibration data and fixed before evaluation. These parameters remained unchanged across all four evaluation stages, while each session received a freshly generated challenge.

### 2.3.2 Data collection procedure

Data collection was organized into two phases. Phase A involved acquiring live bona fide sessions, with captured facial images and videos retained as source material for subsequent prerecorded and manipulated attacks. Phase B involved generating attack sessions while the application presented fresh, independently generated color stimuli. Static-image and prerecorded-video attacks used previously acquired material and did not require a live subject during the attack session. Live deepfake attacks required an operator or driving subject to generate the manipulated stream in real time. All reported attack experiments used virtual-camera delivery.

For each session, the collection procedure recorded the challenge sequence and timestamps, captured frames and timestamps, delivery mechanism, experimental condition, and ground-truth label. These records were used to calculate the session-specific response delay, stimulus–response agreement, and response-duration matching, followed by the final classification according to the decision rule in Section 2.2.8.

The exact sample sizes, device configurations, illumination levels, and degradation settings are reported alongside the corresponding numerical experiments in Section 3.

# 3. Numerical experiments

## 3.1 Method configuration

The study used two subject-disjoint datasets: a separate calibration dataset and an evaluation dataset. The method was calibrated first using the calibration dataset, which contained bona fide sessions only. This dataset was used to select the response-color palette, estimate the reference response delay, compare candidate stimulus durations, train the response-color predictor, and establish the decision thresholds. The evaluation dataset was collected subsequently and used in the main experiments for method validation, generalization, degradation, and evasion analysis. The model, challenge configuration, and decision parameters were fixed before the evaluation data were analyzed and were not further calibrated using the evaluation dataset. At least three sessions were collected per subject. The datasets are summarized in Table 2.

**Table 2.** Summary of the calibration and evaluation datasets.

| Dataset | Subjects | Sessions | Bona fide | Attack |
|---|---|---|---|---|
| Calibration | 6 | 90 | 90 | 0 |
| Evaluation | 15 | 1,128 | 666 | 462 |

Unless otherwise specified, the experiments use the following configuration. Each recorded sequence begins with a fixed neutral preamble consisting of black, white, and gray stimuli. This is followed by six colored stimuli presented in randomized order and ends with a final black stimulus. Thus, each sequence contains four neutral stimuli and six colored stimuli. The neutral preamble is used for baseline acquisition, while all four neutral stimuli are excluded from the randomized challenge.

The active illumination challenge consists of the six colored stimuli. Each of the six colors is presented exactly once, so the color order is sampled without replacement. Each colored stimulus is assigned one duration from the operational stimulus-duration set $\{350, 550, 750\}$ ms. Durations may repeat, but all three durations must appear at least once. The color order and valid duration assignment are generated independently, so no color is systematically associated with a particular duration.

The number of possible colored challenges follows from (1), with six available colors, three duration choices, and six stimuli. There are $6! = 720$ possible color orders and $3^6 - \binom{3}{1}2^6 + \binom{3}{2}1^6 = 540$ valid duration assignments. Therefore, the active challenge has $388{,}800$ possible color-duration sequences. This count excludes the fixed neutral preamble and final black stimulus.

The fixed reference delay $L_{\mathrm{ref}}$ was set to 100 ms based on the calibration results. For session classification, the estimated session-specific response delay was required to fall within 50–150 ms, while the session-level stimulus–response agreement score was required to be at least 0.8. Response-duration matching required all six colored stimuli to satisfy the duration-matching criterion $D_s = 1$. For visualization purposes, a boundary of 0.8 is shown for the normalized duration-matching score, although the actual classification requires complete duration matching. All decision parameters remained unchanged throughout the evaluation experiments.

The rationale for the selected color palette and candidate stimulus durations is provided in Appendices A1 and A2.

## 3.2 Method validation

This section evaluates whether the developed method can distinguish bona fide sessions from the considered attacks under controlled baseline conditions. The baseline bona fide configuration uses an iPhone 15 with Chrome, maximum screen brightness, a stationary subject, and ambient illuminance below 100 lux. Attack sessions use the corresponding injection setups, as described below. The analysis is divided into two parts. First, a disaggregate analysis illustrates the method's behavior for bona fide sessions, prerecorded-video attacks, and real-time deepfake attacks. Second, an aggregate analysis evaluates overall performance across the baseline test sessions.

### 3.2.1 Disaggregate analysis

The disaggregate analysis examines representative sessions to illustrate how the method responds to different stimulus–response characteristics.

*A bona fide session*

Fig. 7 illustrates a subset of frames from a bona fide session. The displayed frames include the neutral stimuli white and gray, followed by cyan, yellow, magenta, green, and red. The corresponding facial response to each stimulus can be observed across the frames, together with a short response delay.

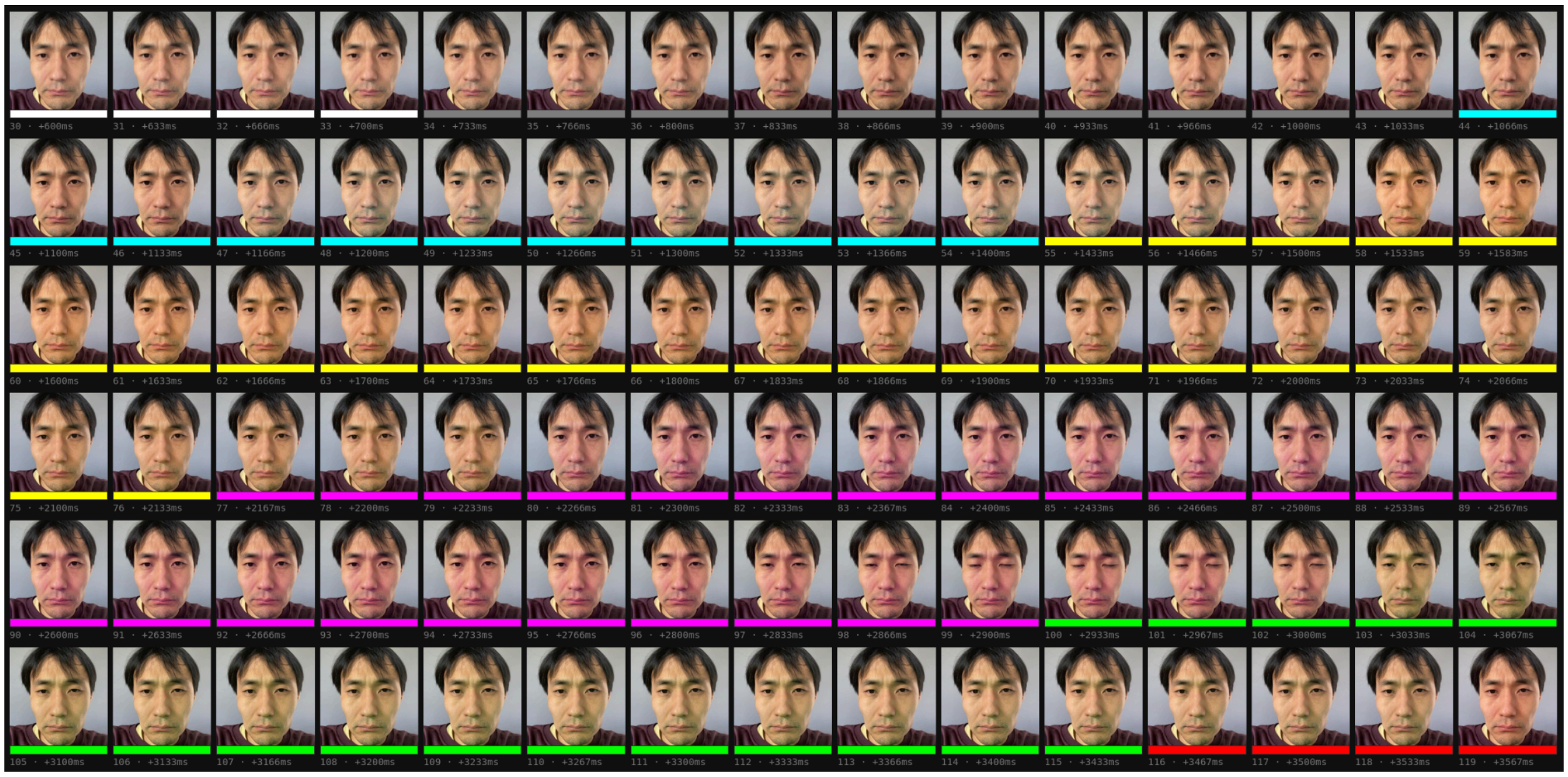

**Figure 7.** Bona fide facial response to the color wave sequence.

Fig. 8 provides a detailed response profile for the same bona fide session by plotting the output of the response-color predictor for each frame. The dominant predicted response colors closely follow the expected stimulus sequence, with a consistent temporal shift throughout the session. For this session, the session-specific response delay is estimated at approximately 100 ms, equivalent to roughly three frames at 30.3 fps.

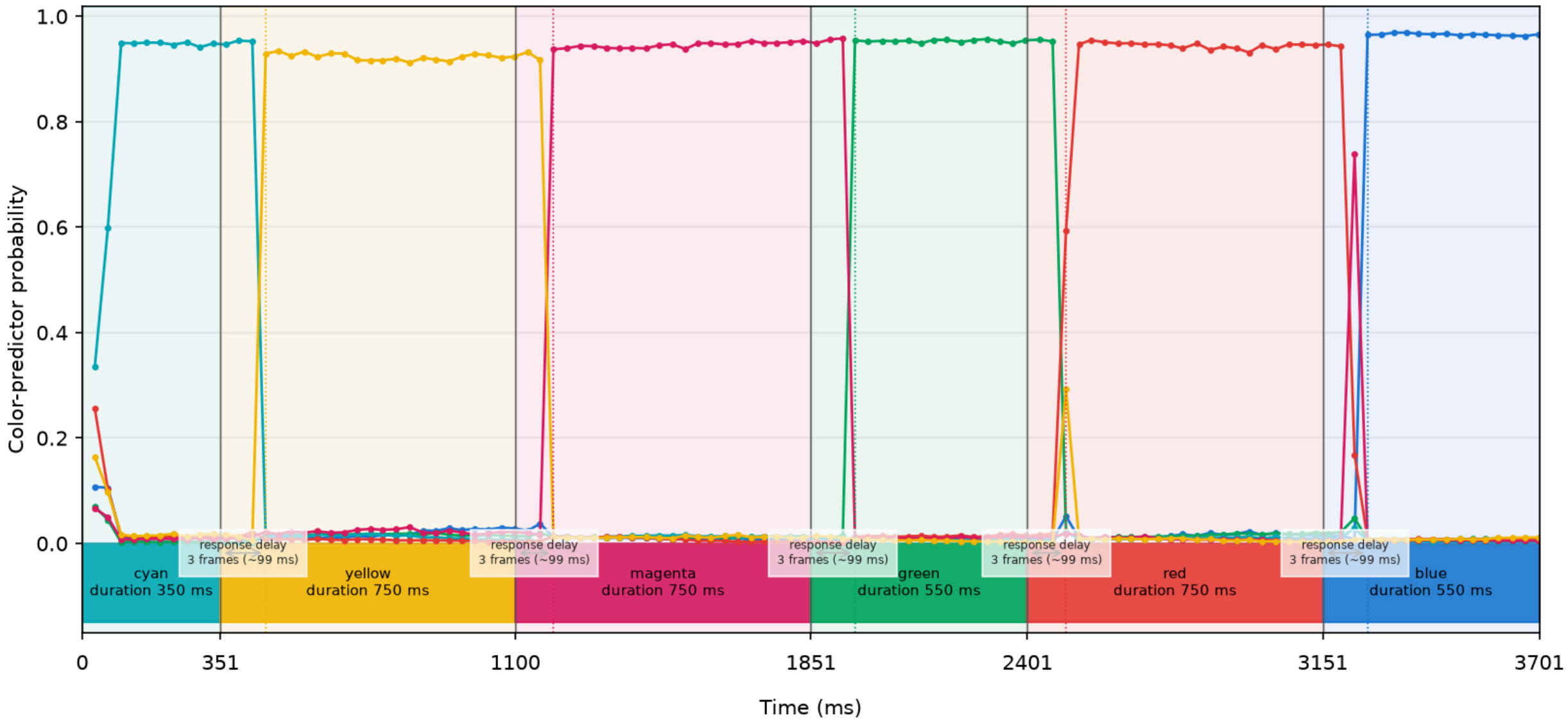

**Figure 8.** Predicted color probabilities for a selected bona fide session.

Using the fixed reference delay $L_{\mathrm{ref}} = 100$ ms, response-duration matching is performed as described in Section 2.2.7. Table 3 summarizes the resulting color-specific response durations. At the measured frame rate of 30.3 fps, each frame corresponds to approximately 33 ms; thus, about 10, 16–17, and 22–23 matched frames correspond to the nominal durations of 350, 550, and 750 ms, respectively.

**Table 3.** Frame-based estimates of color-specific response durations for a bona fide session.

| Stimulus | Nominal duration (ms) | Delay-shifted interval frames | Matched frames | Matched-frame estimate (ms) | Mapped response duration (ms) |
|---|---|---|---|---|---|
| Cyan | 350 | 10 | 10 | 330 | 350 |
| Yellow | 750 | 23 | 22 | 726 | 750 |
| Magenta | 750 | 22 | 22 | 726 | 750 |
| Green | 550 | 17 | 16 | 528 | 550 |
| Red | 750 | 22 | 21 | 693 | 750 |
| Blue | 550 | 17 | 16 | 528 | 550 |

The session shows a clear and timely color-correlated facial response, with an estimated session-specific delay of approximately 100 ms, within the acceptable interval of 50–150 ms. The response-color predictions closely follow the expected stimulus sequence, and all six stimuli satisfy the duration-matching criterion $D_s = 6$. This behavior contrasts with the delayed or misaligned responses observed in the attack sessions presented below.

*A pre-recorded video attack session*

Fig. 9 illustrates a subset of frames from a prerecorded-video attack session, arranged similarly to the bona fide example. Unlike the bona fide case, the facial appearance remains largely unchanged across the displayed stimuli, with no clearly observable stimulus-dependent color response.

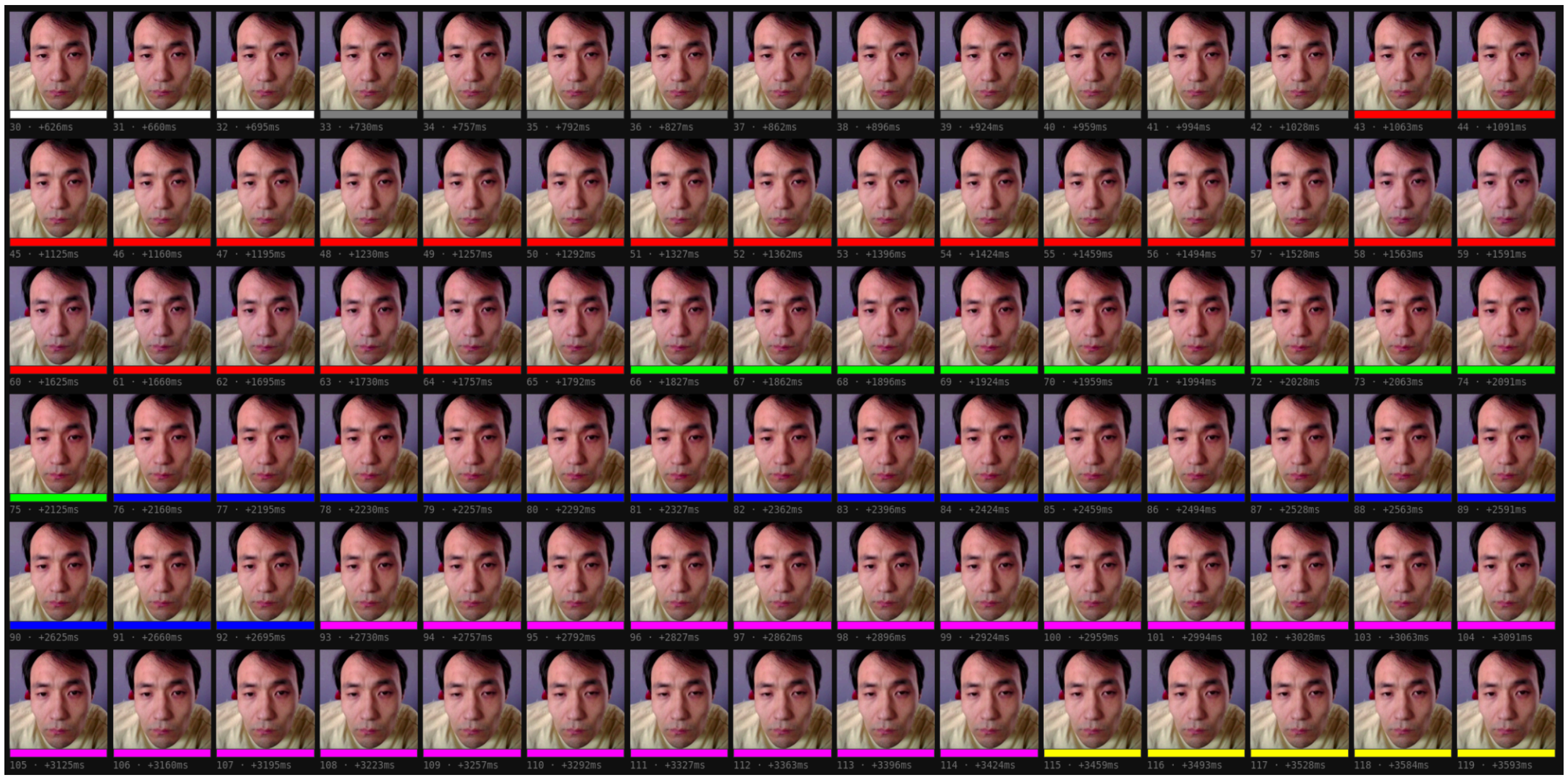

**Figure 9.** Pre-recorded video attack facial response to the color wave sequence.

Fig. 10 shows the response profile for the same attack session. Unlike the bona fide case, no dominant predicted response colors follow the expected stimulus sequence. The session presents red for 750 ms, green for 350 ms, blue for 550 ms, followed by magenta, yellow, and cyan, each for 750 ms. However, the predicted response remains largely unchanged: magenta dominates throughout the recording, with red generally the second-highest prediction, while the remaining colors stay near zero. As no clear stimulus-specific response pattern is observed, the session-specific response delay cannot be reliably estimated.

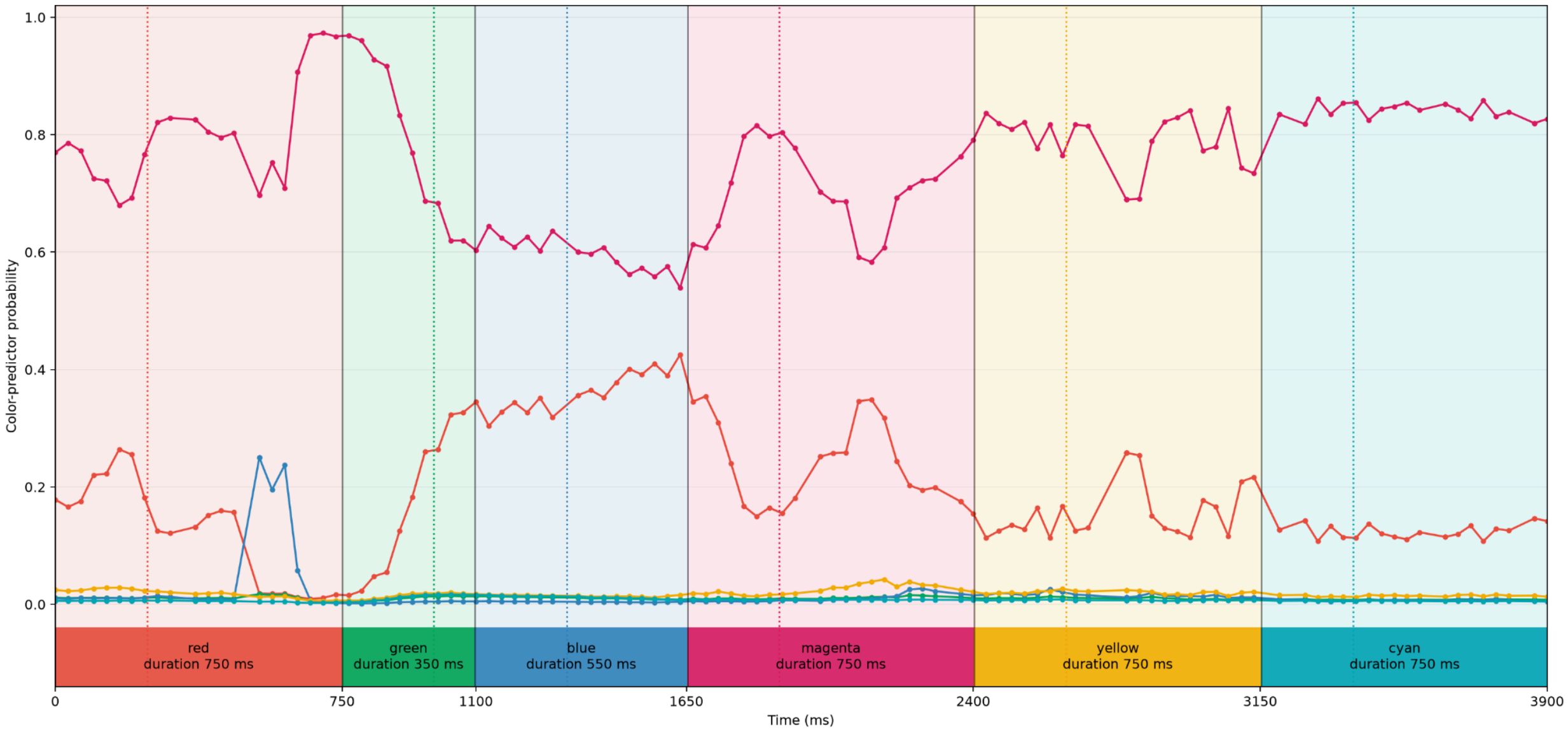

**Figure 10.** Predicted color probabilities for a selected pre-recorded video attack session.

This example illustrates how prerecorded-video attacks may be distinguished from bona fide sessions through the absence of a stimulus-aligned facial response. Unlike the genuine session, the injected video does not reproduce the current challenge sequence, resulting in inconsistent response-color predictions and preventing reliable temporal alignment.

### *A live deepfake attack session*

Fig. 11 illustrates a subset of frames from a live deepfake (face-swap) attack session. The attack was generated using DeepFaceLive[2], with a subject positioned in front of a Logitech C270 HD webcam and the subject's face replaced in real time. The processed stream was routed through OBS Studio's virtual camera and used as input to the color wave method running in Google Chrome on the same PC. The measured frame rate was 30.03 fps.

Unlike the prerecorded-video attack, the live deepfake exhibits a visible response to the color stimuli. However, the response is substantially more delayed than in a bona fide session. For example, the response to the red stimulus is shifted by more than 25 frames, corresponding to over 0.8 s at 30.03 fps. This exceeds the maximum stimulus duration of 750 ms and is substantially larger than the delay observed for bona fide sessions.

[2] https://github.com/iperov/DeepFacelive

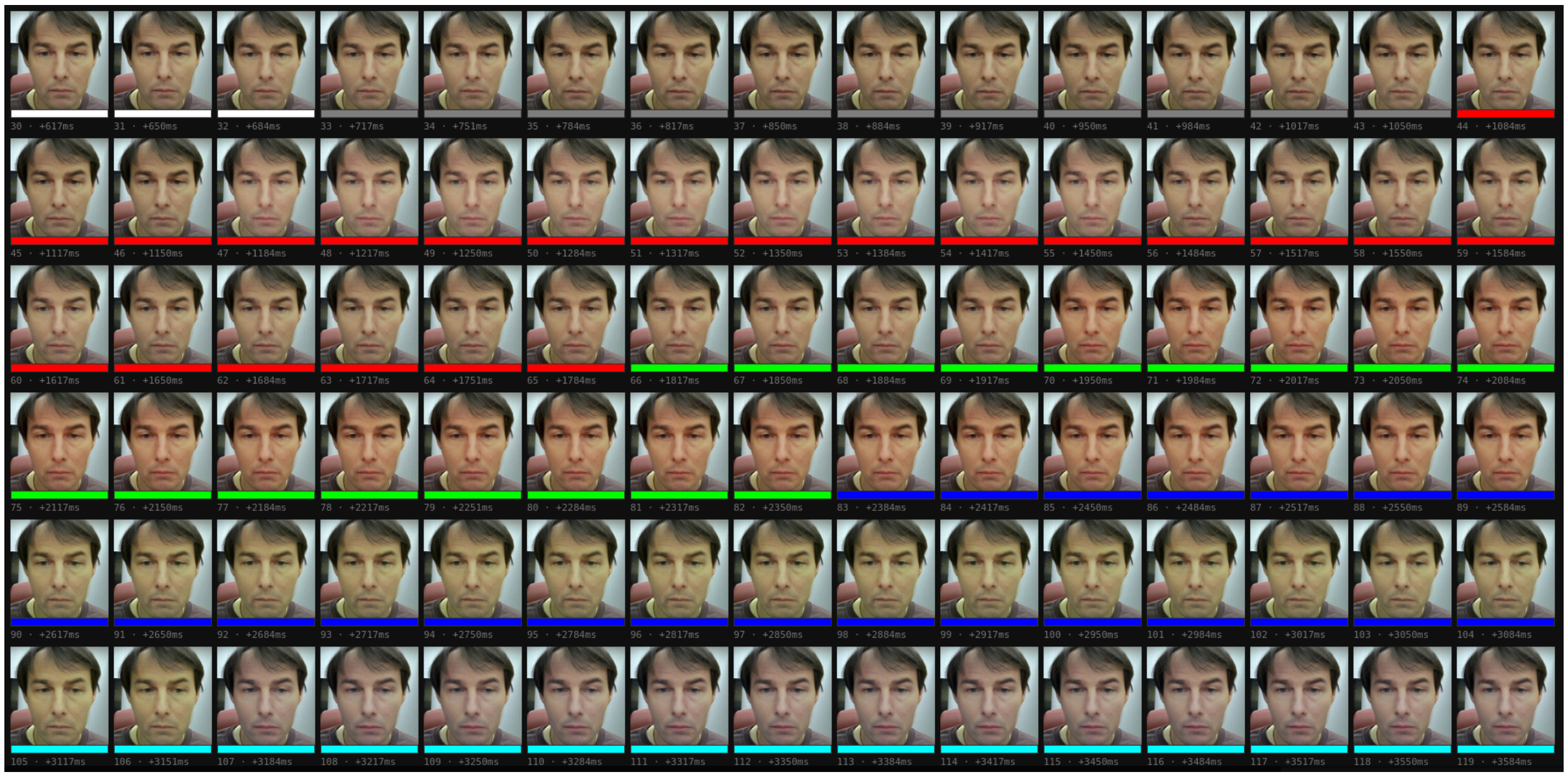

**Figure 11.** Live deepfake attack facial response to the color wave sequence.

As illustrated in Fig. 12, the live deepfake session exhibits a clear color-correlated facial response, but with an observed temporal shift exceeding 750 ms, beyond the delay-estimation range specified in Section 2.2.4.

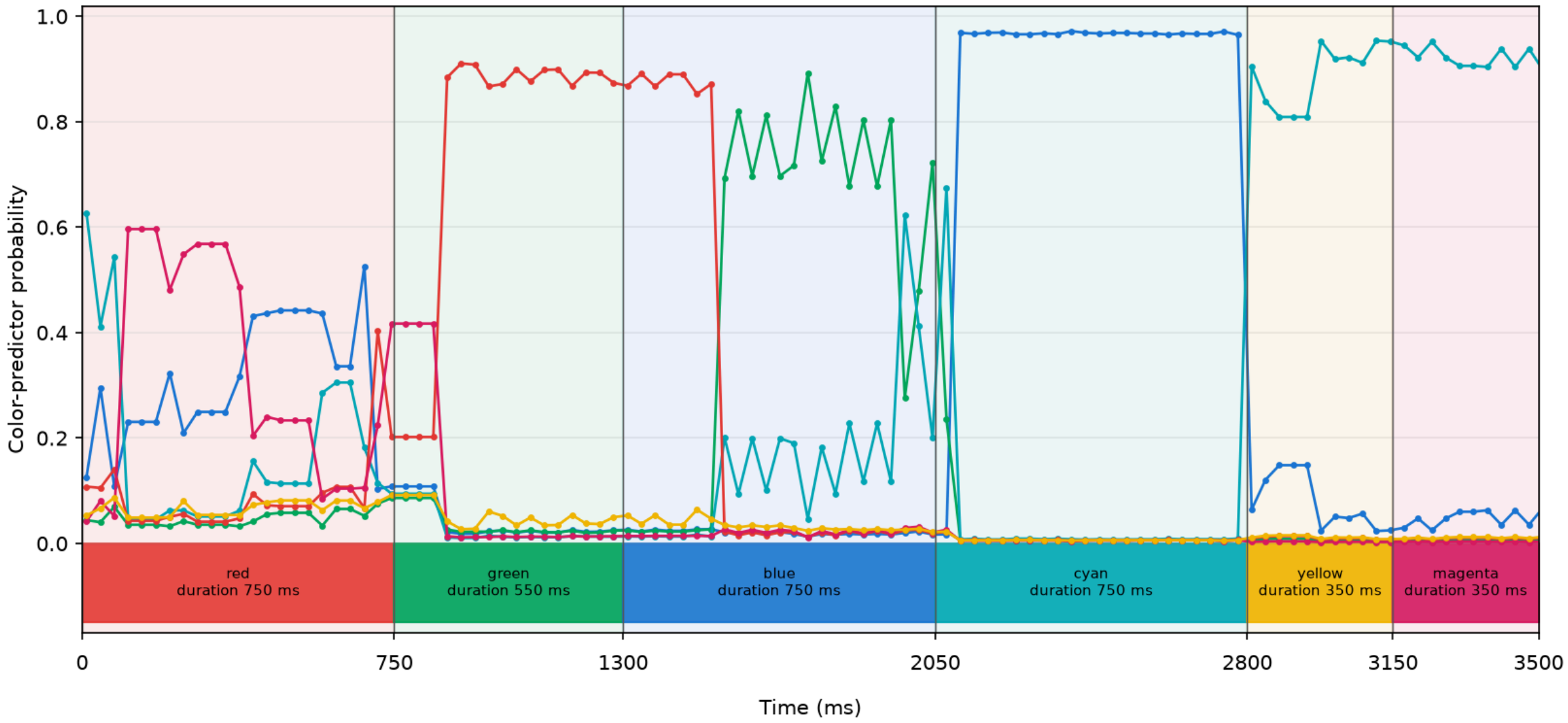

**Figure 12**. Predicted color probabilities for a selected live deepfake attack session.

The experiments were conducted on two computers: a desktop equipped with an Intel Core i7-8700 CPU, NVIDIA GeForce GTX 1080 Ti GPU (11 GB), and 32 GB RAM; and an ASUS TUF Gaming A15 laptop equipped with an AMD Ryzen 7 7435HS CPU, NVIDIA GeForce RTX 3050 GPU (4 GB), and 16 GB RAM. The same qualitative behavior was observed on both systems. This example demonstrates that a live deepfake can preserve a stimulus-correlated facial color response while introducing substantial temporal delays. Unlike the prerecorded-video example, the observed inconsistency is primarily temporal rather than an absence of color response, highlighting the complementary role of response-delay analysis in injection attack detection.

### 3.2.2 Aggregate analysis

The aggregate analysis evaluates the classification performance of the developed method under controlled baseline conditions. The data were collected following the procedure in Section 2.3.2 and comprise 49 bona fide, 84 prerecorded-video, and 80 live-deepfake sessions, totaling 213 sessions. Each session is classified according to the decision rule in Section 2.2.8, which requires all three criteria—response delay, session-level agreement, and response-duration matching—to be satisfied for bona fide classification. Failure of any criterion results in an attack classification.

Fig. 13 summarizes the results. Under baseline conditions, the method correctly classifies all 213 sessions, achieving 100% accuracy. This demonstrates that the combined decision criteria can distinguish bona fide and attack sessions under the tested baseline conditions. However, this result alone does not establish the method's generalization or robustness. The subsequent experiments therefore evaluate whether the observed classification performance persists under new conditions, signal degradation, and deliberate evasion attempts.

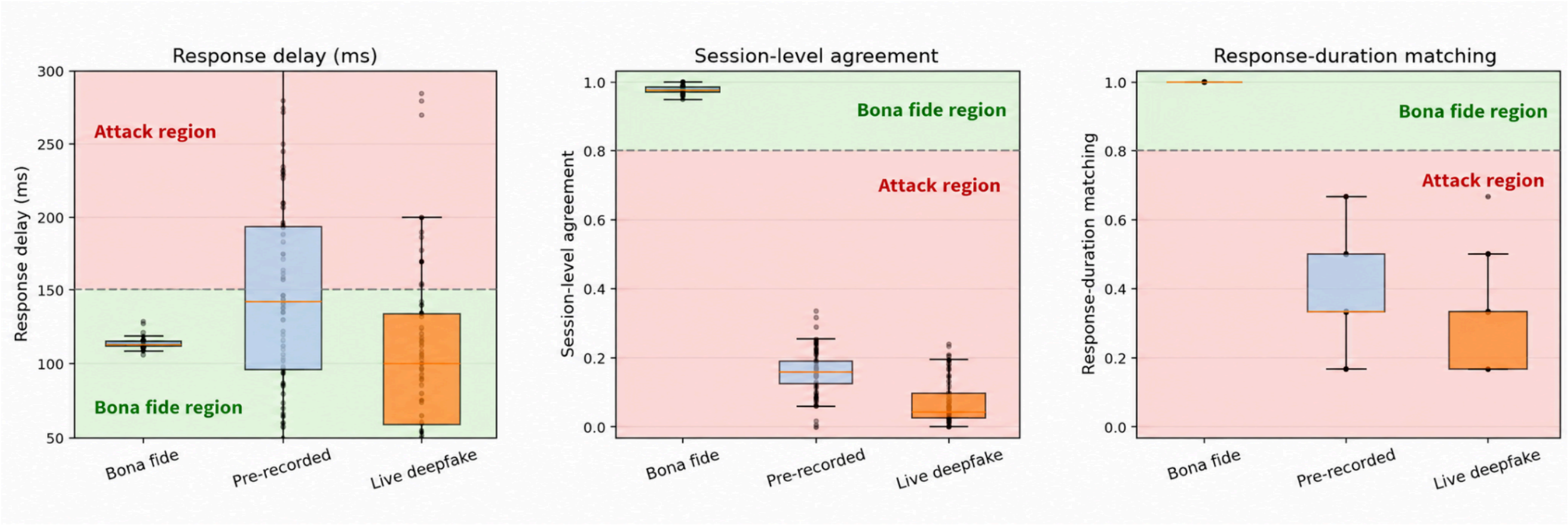

**Figure 13.** Distribution of the three decision criteria under baseline conditions, with bona fide and attack decision regions.

## 3.3 Generalization

This section assesses generalization across subjects, devices, browsers, backgrounds, motion, lighting, and attack implementations, all of which can potentially influence the captured stimulus–response signal. We first evaluate bona fide-session classification under these varying conditions, followed by attack classification across different environmental conditions and injection implementations.

### 3.3.1 Analysis of bona fide sessions

The bona fide generalization experiment used a reference condition consisting of an iPhone 15, Chrome, an office-desk background, a stationary subject, and natural lighting. Fifteen participants completed the color wave challenge under this condition, after which we varied one factor at a time while keeping the others fixed. We evaluated multiple Apple and Android smartphones and a PC; Chrome, Firefox, and Safari; office, kitchen, and library backgrounds; stationary and moving subjects; and natural and artificial lighting. For the motion condition, participants moved around the office while completing the challenge. Each participant completed at least three sessions per scenario.

Fig. 14 summarizes response delay, session-level agreement, and response-duration matching, which are combined using an AND rule: a session is classified as bona fide only if its delay is within 50–150 ms, agreement is at least 0.8, and all active stimuli have matching response durations; otherwise, it is classified as an attack. The boxplots summarize session distributions, while individual points represent individual sessions.

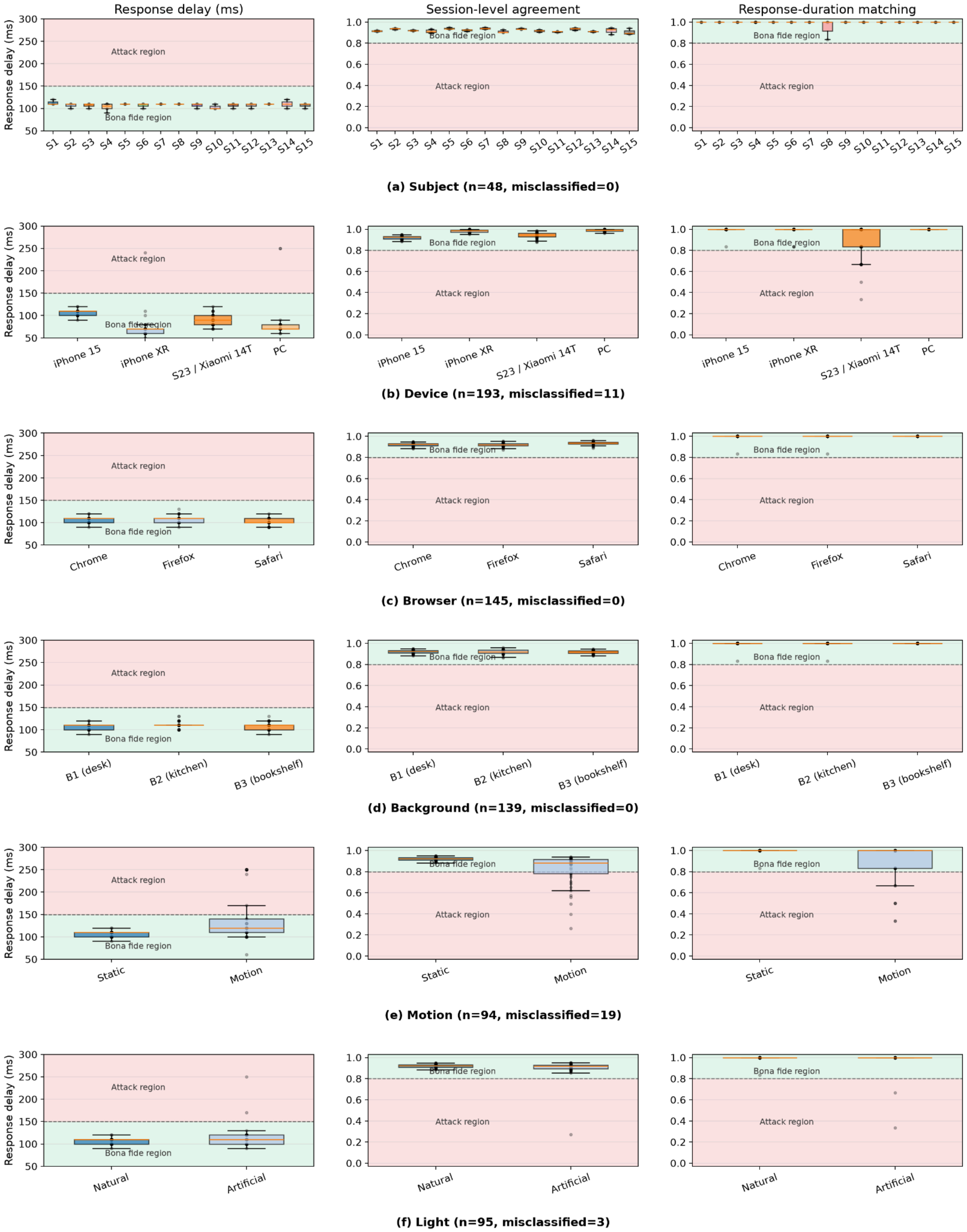

**Figure 14.** Bona fide-session generalization evaluation across (a) subjects, (b) devices, (c) browsers, (d) backgrounds, (e) motion, and (f) lighting.

As shown in Fig. 14, most conditions produced similar results. Misclassifications occurred mainly for some Android sessions, where dropped frames distorted duration estimates, and during motion, where spatial misalignment degraded agreement and duration matching. Overall, bona fide classification was robust across the evaluated subjects, browsers, backgrounds, lighting conditions, and most devices, but remained sensitive to dropped frames and subject motion.

### 3.3.2 Analysis of attacks

Attack generalization was evaluated using prerecorded-video and live face-swapping attacks. Static-image attacks were treated as a special case of prerecorded-video attacks, in which the same frame is repeated. We evaluated the same environmental factors as for bona fide sessions. Additionally, we examined three virtual-camera applications—OBS Studio, SplitCam, and ManyCam—and two face-swapping implementations in DeepFaceLive: DFM, which uses a trained face model, and Insight, which performs face swapping from a single reference image.

The tested environmental factors and virtual-camera applications showed negligible differences in attack-classification outcomes; therefore, these results are not presented separately. The more pronounced differences were observed between the face-swapping implementations, as summarized in Fig. 15.

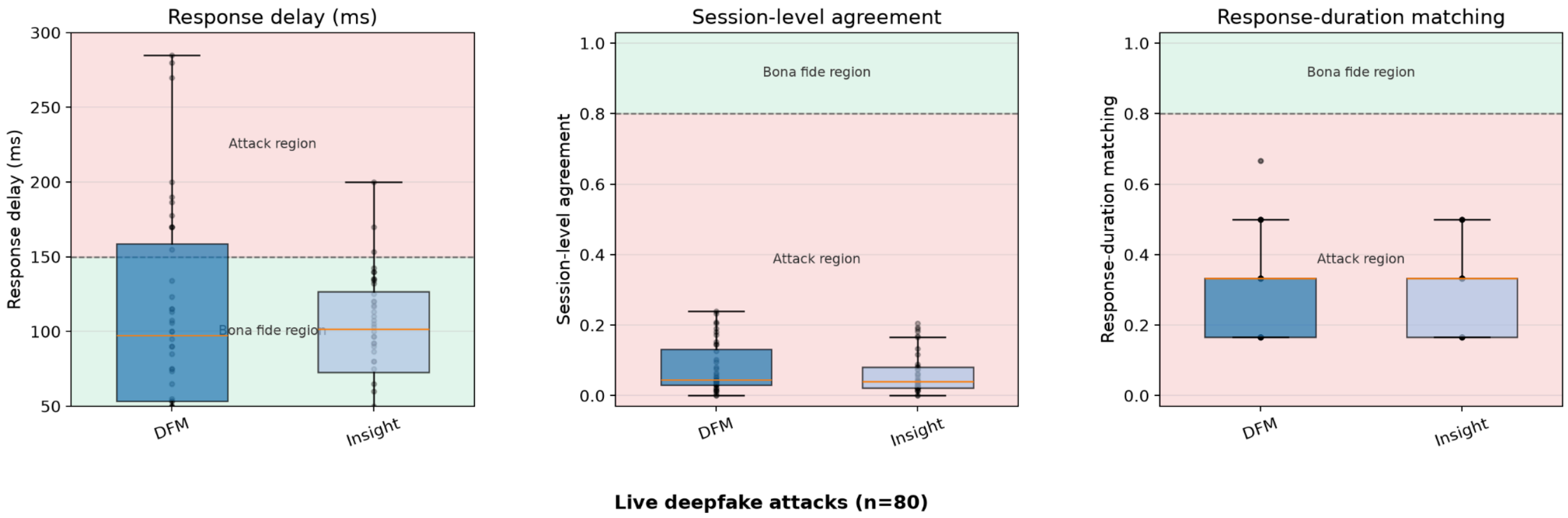

**Figure 15.** Generalization across live face-swapping implementations.

As shown in Fig. 15, sessions generated using both face-swapping implementations were classified as attacks. Their processing delays were associated with reduced stimulus–response agreement and mismatched response durations, as expected when responses are evaluated using the fixed reference delay. Overall, attack classification remained stable across the tested environmental and virtual-camera variations and generalized to both DFM and Insight.

## 3.4 Robustness to degradation

This section evaluates robustness to degradation in screen brightness, ambient illumination, spatial resolution, and frame rate. Because the observed impact of these degradations primarily concerns the false rejection of bona fide sessions, the analysis focuses on bona fide data and examines their effects on classification performance. The evaluated degradations may reduce the bona fide acceptance rate, whereas no corresponding increase in missed attacks was observed in the available evaluation. Fig. 16 and Fig. 17 summarize the results.

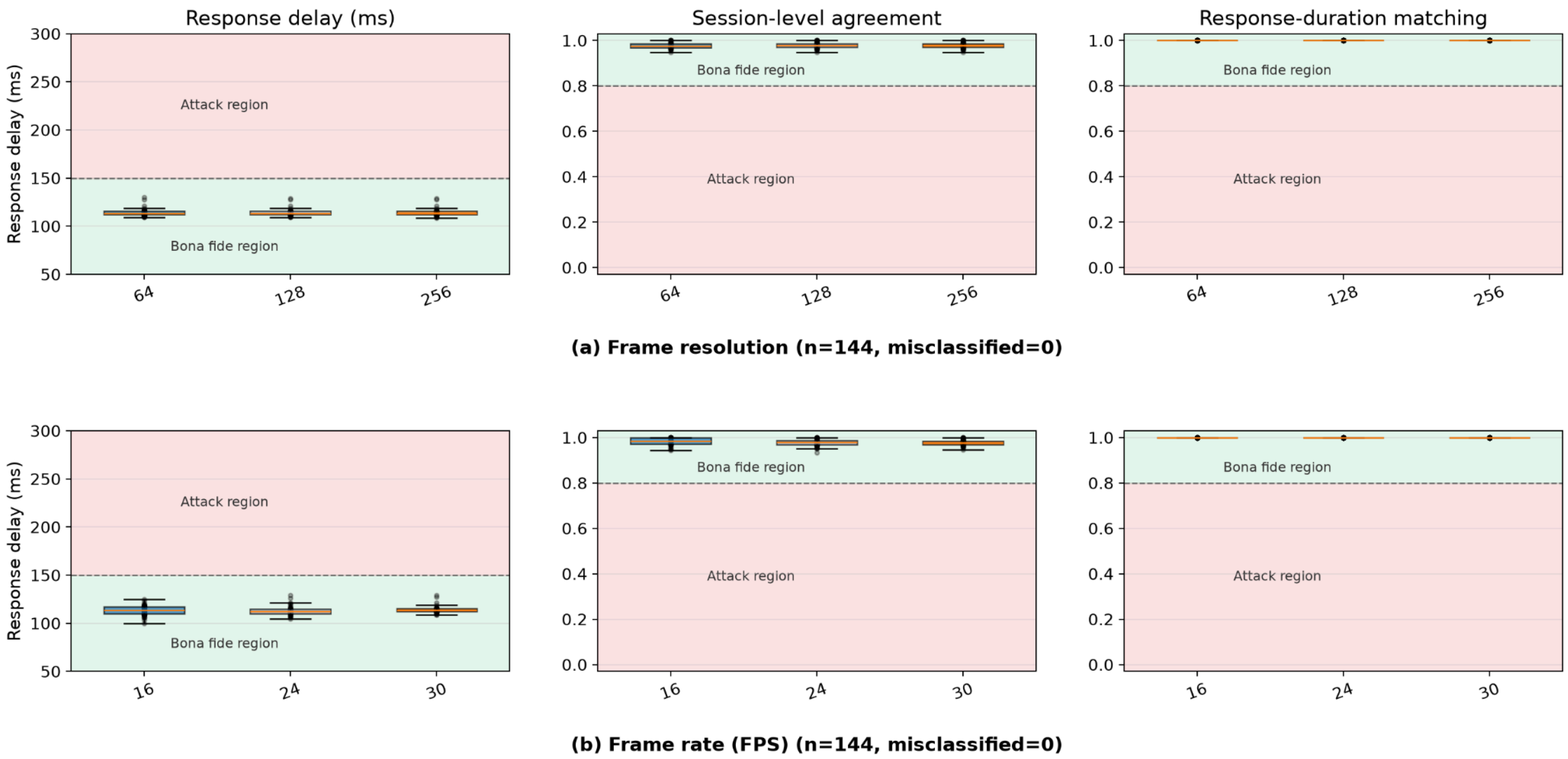

**Figure 16.** Frame resolution and frame-rate degradation robustness for bona fide sessions.

As shown in Fig. 16, resolution and frame-rate degradation had no noticeable effect on bona fide classification performance within the tested ranges. This finding does not exclude the effects of irregular frame loss observed in Section 3.3.1.

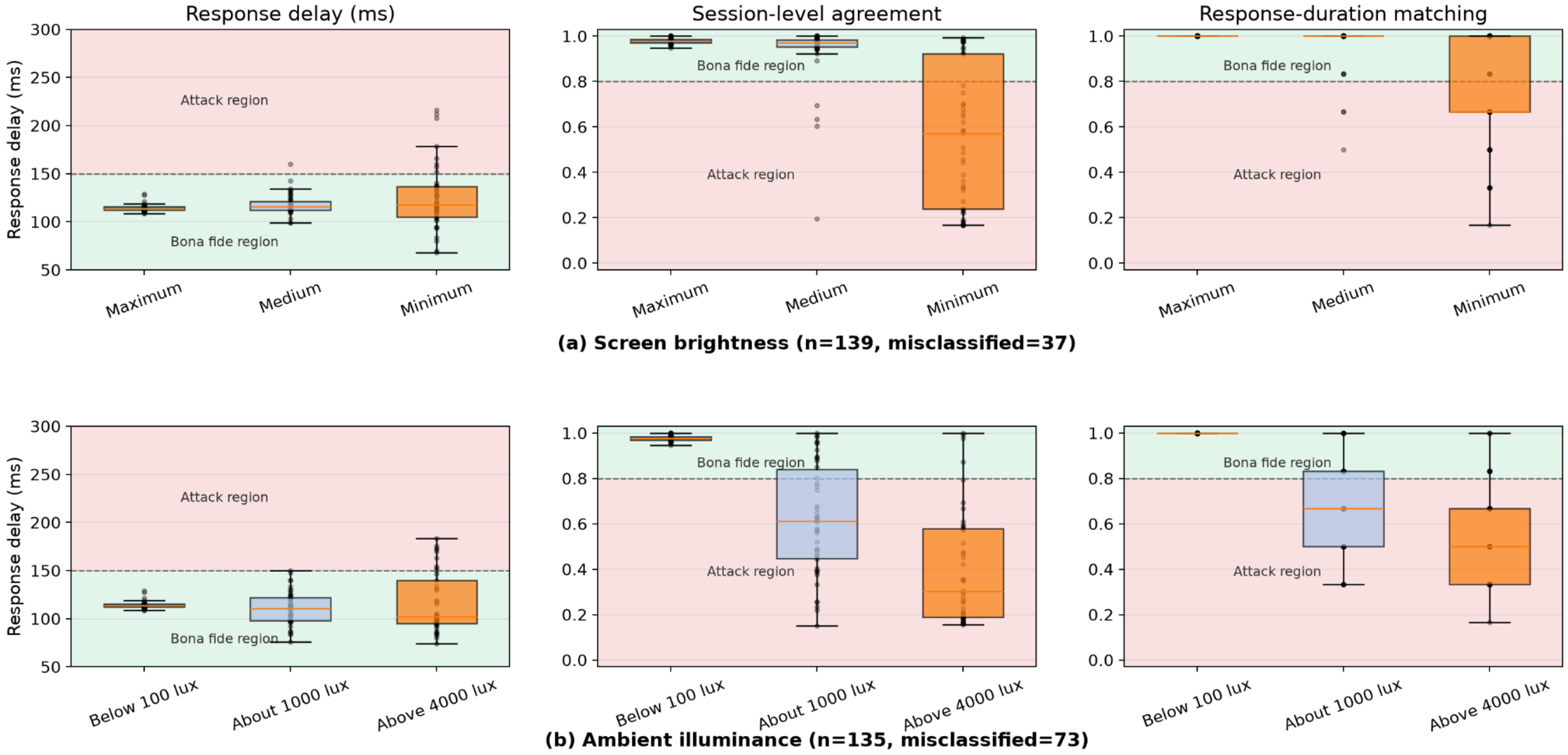

**Figure 17.** Screen brightness and ambient illuminance degradation robustness for bona fide sessions.

As shown in Fig. 17a, screen brightness has a substantial effect on performance, with higher brightness levels producing better results. This trend is likely because a brighter screen illuminates the face more strongly, producing a more pronounced stimulus–response signal. At minimum brightness, the screen casts little light onto the face, resulting in a weaker or barely detectable response. Excessively intense ambient illumination also degrades classification performance because it reduces the contrast between stimulus-induced facial color changes and background illumination, effectively washing out the color response (Fig. 17b).

## 3.5 Resistance to evasion

This section evaluates whether attacks constructed from prerecorded genuine color responses can reproduce the correct response colors with the timing required by a fresh randomized challenge. Two evasion strategies are considered: accelerated replay and stimulus-triggered frame switching or recoloring.

### 3.5.1 Accelerated replay

The first evasion strategy evaluates accelerated replay as a potential attack. In this scenario, prerecorded video containing a genuine response to color stimuli is played back at different speeds, causing responses to multiple colors to potentially appear within the duration of a single stimulus. We consider 1×, 2×, 3×, and 10× playback speeds.

At 1× speed, the scenario represents an attacker replaying a bona fide user's video at its original speed. At higher acceleration factors, individual colors may occasionally coincide with the current stimuli, but the replayed sequence is not synchronized with the fresh challenge. Acceleration compresses the temporal sequence of recorded responses without adapting their order to the current stimuli, making isolated matches insufficient to pass the verification reliably. Fig. 18 summarizes the results.

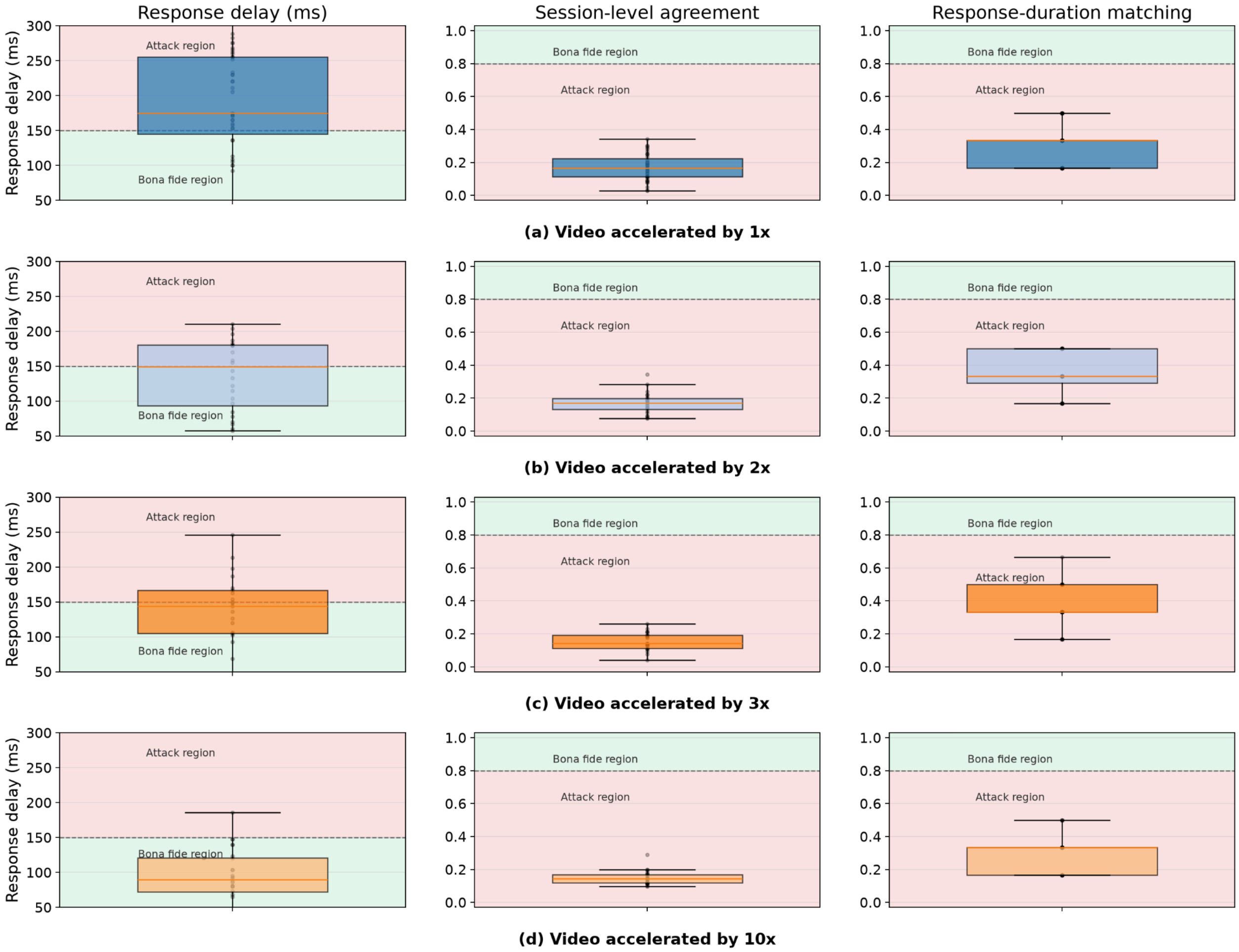

**Figure 18.** Video-injection attacks under different playback acceleration factors.

As shown in Fig. 18, all tested replay attacks were correctly classified as attacks, indicating that playback acceleration alone was insufficient to reproduce the required stimulus–response correspondence.

### 3.5.2 Stimulus-triggered frame switching

The second evasion strategy comprises two approaches: manual and automatic stimulus-conditioned video manipulation. In the manual approach, the attacker configures hotkeys in the virtual-camera software to select video frames based on the active stimulus. For example, when the service displays a red stimulus, the attacker presses a hotkey to switch the camera feed to a frame containing a red response or to apply a red overlay to the current frames. In the automatic approach, the stimulus color is detected programmatically, and the virtual camera switches to the corresponding frame or applies the matching color without user input. Fig. 19 summarizes the results for both manual and automatic manipulation.

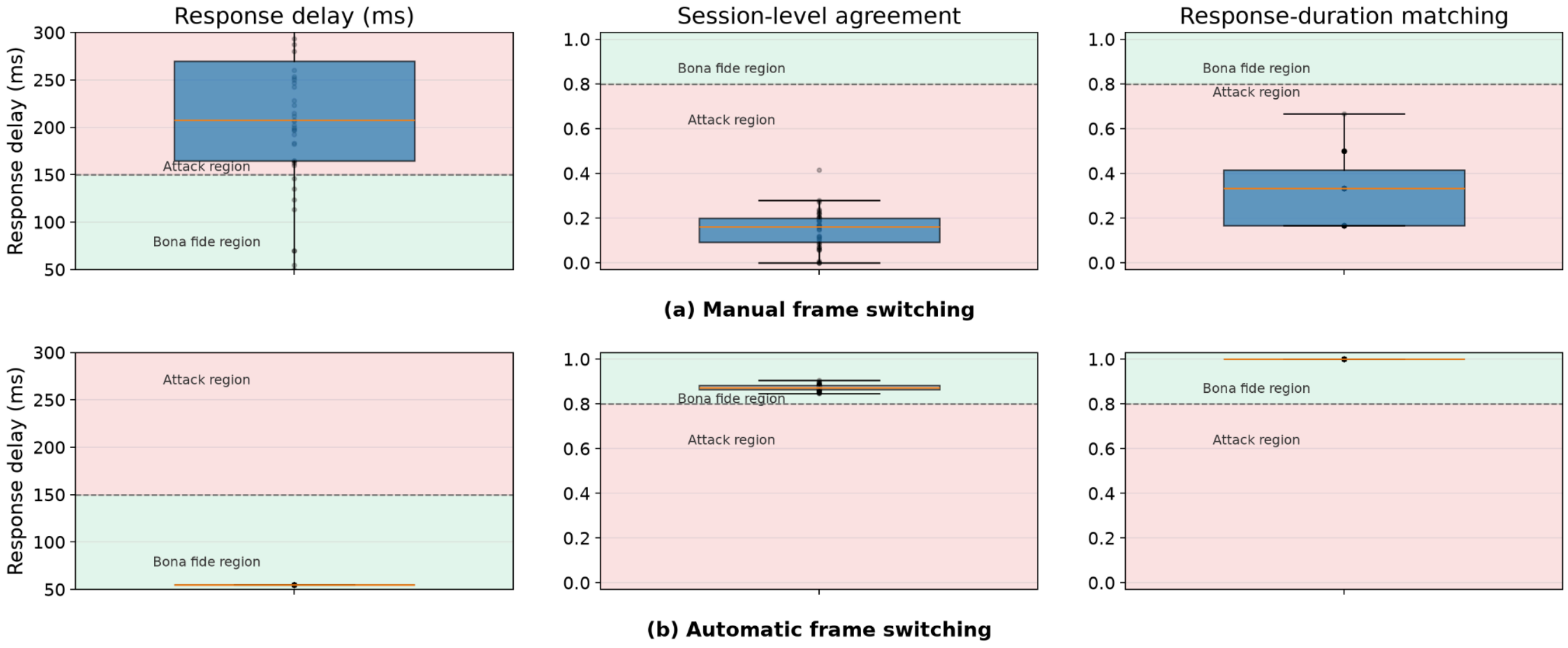

**Figure 19**. Manual and automatic stimulus-conditioned video-injection attacks.

As shown in Fig. 19, the manual strategy was detected, most likely because observing the stimulus, pressing a hotkey, and switching or recoloring the camera feed introduced a measurable delay that disrupted temporal synchronization. In contrast, automatic manipulation demonstrated a potential vulnerability to stimulus-conditioned evasion. By eliminating manual intervention, the attacker can respond more rapidly to the displayed stimulus and potentially reproduce the expected color changes within the acceptable response-delay interval. This suggests that sufficiently responsive automated attacks may satisfy the temporal and color-agreement criteria without relying on genuine facial illumination responses.

## 3.6 Ablation study of the decision criteria

This experiment evaluates the individual and combined contributions of the three decision criteria: response delay, session-level stimulus–response agreement, and response-duration matching. For every session, criterion values were converted into normalized attack scores, where larger values indicate stronger evidence of an attack: response delays outside the acceptable interval, lower stimulus–response agreement, and fewer duration-matched stimuli. Combined scores use the maximum normalized violation across the selected criteria. Unlike the fixed-threshold classification in Section 2.2.8, this analysis evaluates discrimination across varying score thresholds using receiver operating characteristic (ROC) curves.

The evaluation uses all root-level sessions from the method-validation, generalization, degradation, and evasion stages. Calibration sessions and the separate FPS and resolution experiments are excluded. Results are summarized in Fig. 20.

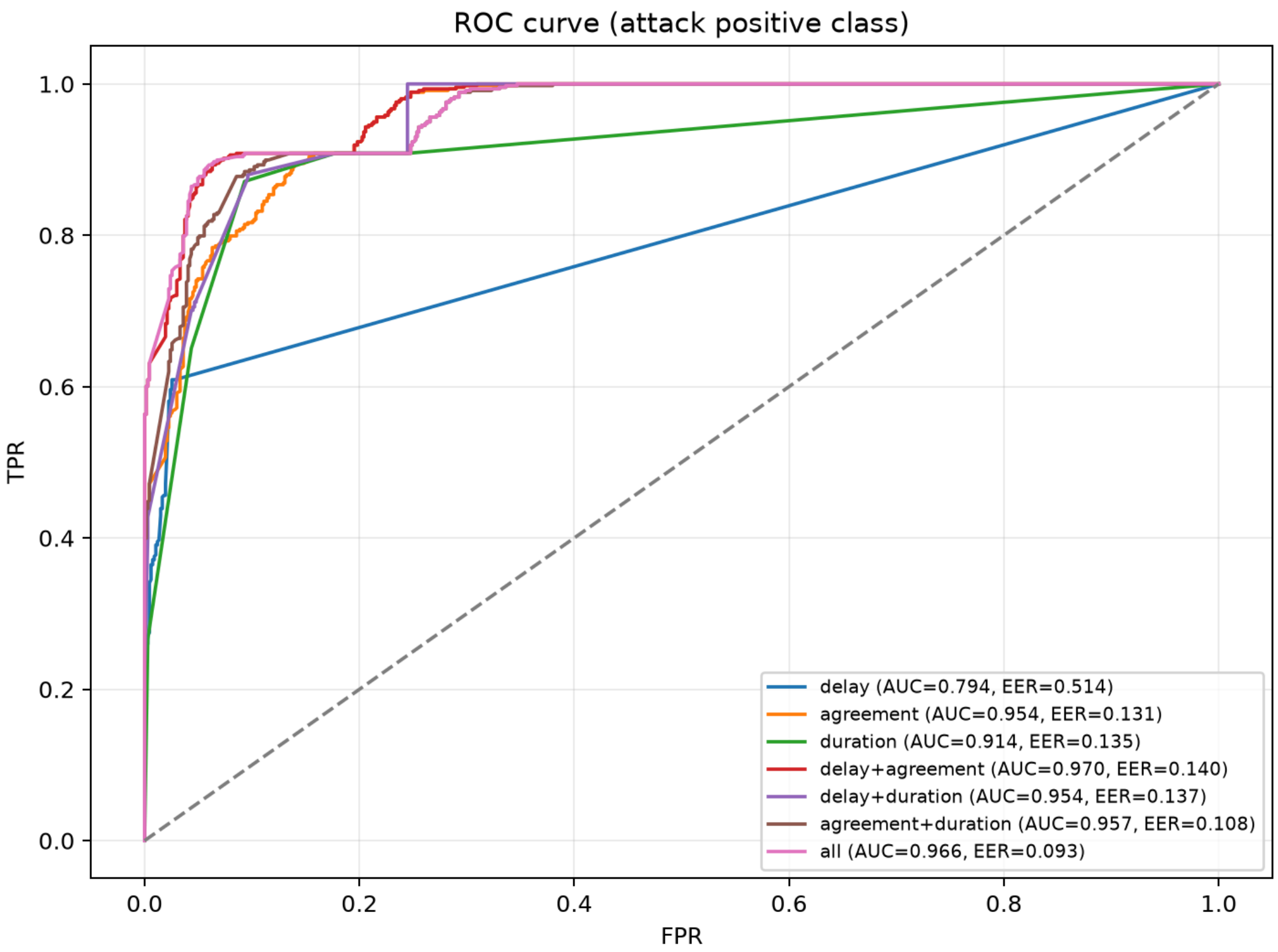

**Figure 20.** ROC curves for individual and combined attack criteria.

Among the individual criteria, stimulus–response agreement achieved the highest ROC-AUC of 0.954 and the lowest equal error rate (EER) of 0.131. Duration matching achieved a ROC-AUC of 0.914 and EER of 0.135, while response delay alone showed weaker discrimination, with a ROC-AUC of 0.794 and EER of 0.514.

Combining criteria generally improved discrimination. The delay-plus-agreement combination achieved the highest ROC-AUC (0.970), while combining all three criteria produced the lowest EER (0.093) with a ROC-AUC of 0.966. These results suggest that stimulus–response agreement provides the strongest individual discriminatory signal, while duration matching and response delay contribute complementary information when combined. The three-criterion combination achieves the lowest error rate at the equal-error operating point, whereas delay plus agreement provides the strongest overall ranking performance.

## 3.7 Discussion

The main finding is that the color wave method provides useful complementary evidence for biometric data injection detection in the tested virtual-camera setting. Its signal is physically grounded: bona fide camera input carries a screen-induced facial response linked to the current challenge, whereas the evaluated injected streams generally could not reproduce that response with the same timing. The ablation analysis further indicates that stimulus–response agreement provides the strongest individual discriminatory signal, while the other criteria contribute complementary information when combined. These findings support the feasibility of the method and the value of its transparent, injection-focused evaluation. However, they do not establish deployment-level performance or comprehensive resistance to adaptive attacks.

The main operational limitations concern acquisition conditions and response reliability. The baseline-referenced predictor is vulnerable to movement and spatial misalignment, while dropped frames can make a valid response appear too short for reliable duration estimation. The optical signal also depends on screen brightness and ambient illumination; a weak or washed-out response can produce a bona fide rejection. These limitations suggest acquisition-aware remedies: positioning guidance and face alignment, a later or periodically refreshed neutral reference, timestamp-based frame handling, and retry or reacquisition when a response window is undersampled. Brightness and ambient-light checks, user guidance, and adaptive brightness or duration are also plausible, but should be validated without simply weakening the decision rule. The proposed remedies would introduce usability and security trade-offs and were not demonstrated by the present experiments.

The most important security limitation is the vulnerability to automatic stimulus-conditioned manipulation identified in the evasion experiments. Randomized color order and duration make prerecorded replay difficult, but they do not prevent an attacker from observing the current challenge and modifying an injected stream in real time. A promising extension is to combine randomized multi-color illumination with spatially structured or region-specific patterns, making simple frame recoloring or frame switching less effective. *Tang et al. (2018)* analyze reflected light and facial structure from designed image stimuli, while *Guo et al. (2023)* use a distinct probing pattern and its corneal reflection. These precedents support the direction, but do not establish that a combined-color or patterned challenge would defeat the tested bypass; that claim requires new adaptive-attack experiments.

The method should therefore be viewed as one layer in a broader authentication system. It assesses whether the observed facial response is consistent with the presented screen challenge, but cannot independently establish that the response originates from genuine physical facial illumination. It also does not verify identity and should not replace face recognition, presentation attack detection, or stream-integrity controls. Future work should evaluate the proposed challenge extensions and the existing method against direct application-level injection and broader low-latency attacks, while measuring their effects on bona fide rejection, user burden, and environmental variability.

# 4. Conclusion

This study addressed three objectives. First, it established an injection-specific threat model and a reproducible implementation of active screen illumination for remote face authentication, contributing a systematic empirical evaluation rather than a new illumination principle. Second, it examined whether facial-response color, timing, and duration distinguish bona fide input from injected biometric video. The results demonstrated strong discrimination under controlled conditions and evaluated generalization across subjects, devices, environments, and attack implementations. Third, the study characterized the method's operating limits. Subject movement, dropped frames, low screen brightness, and strong ambient illumination increased the risk of bona fide rejection. Moreover, the successful automatic stimulus-conditioned evasion demonstrated that randomized challenges alone cannot prevent sufficiently responsive adaptive attacks. These findings support the method's feasibility as a complementary injection-detection mechanism, rather than standalone protection.

The evaluation focused on virtual-camera injection using selected replay, face-swap, and stimulus-conditioned attacks. Direct application-level injection, broader adaptive strategies, and deployment-scale populations were not examined. Furthermore, the proposed remedies for acquisition and illumination limitations remain unvalidated. The findings therefore characterize performance under the tested conditions without establishing general attack resistance or deployment-level reliability.

Future work should improve acquisition reliability through better face alignment, reference-frame selection, and capture-quality assessment while examining the associated usability and security trade-offs. Spatially structured and multi-color challenges warrant investigation as potential defenses against adaptive manipulation. Further studies should evaluate direct injection, broader low-latency attacks, and complementary stream-integrity controls using larger and more diverse populations. Ultimately, the method should be assessed as one component of a layered biometric authentication system.

# Acknowledgements

The authors used ChatGPT (OpenAI) to assist with grammar and language polishing of the manuscript. All generated edits were reviewed, verified, and finalized by the authors.

# References

Agarwal, A., & Ratha, N. (2025). Detection of identity swapping attacks in low-resolution image settings. *Journal of Information Security and Applications*, 89, 103911. https://doi.org/10.1016/j.jisa.2024.103911

Amazon Web Services. (2023, November 27). Amazon Rekognition Face Liveness. https://docs.aws.amazon.com/ai/responsible-ai/rekognition-face-liveness/overview.html

Bud, A. (2018). Online pseudonym verification and identity validation (U.S. Patent No. 10,133,943 B2). U.S. Patent and Trademark Office. https://patents.google.com/patent/US10133943B2/en

Bud, A., & Newell, A. (2025). Biometric methods for online user authentication (U.S. Patent No. 12,192,198 B2). U.S. Patent and Trademark Office. https://patents.google.com/patent/US12192198B2/en

Carta, K., Barral, C., El Mrabet, N., & Mouille, S. (2022). Video injection attacks on remote digital identity verification solution using face recognition. In Proceedings of the 13th International Multi-Conference on Complexity, Informatics and Cybernetics: IMCIC 2022 (Vol. 2, pp. 92-97). International Institute of Informatics and Cybernetics. https://doi.org/10.54808/IMCIC2022.02.92

Carta, K., Huynh, A., Mouille, S., El Mrabet, N., Barral, C., & Brangoulo, S. (2023). How video injection attacks can even challenge state-of-the-art face presentation attack detection systems. In Proceedings of the 14th International Multi-Conference on Complexity, Informatics and Cybernetics: IMCIC 2023 (pp. 105-112). International Institute of Informatics and Cybernetics. https://doi.org/10.54808/IMCIC2023.01.105

Carta, K., Huynh, A., Mouille, S., & El Mrabet, N. (2024). An overview of biometric data injection attacks on remote identity proofing solutions [Preprint]. SSRN. https://ssrn.com/abstract=4855561

Castelblanco, A., Rivera, E., Solano, J., Tengana, L., López, C., & Ochoa, M. (2022). Dynamic face authentication systems: Deep learning verification for camera close-up and head rotation paradigms. *Computers & Security*, 115, 102629. https://doi.org/10.1016/j.cose.2022.102629

Chan, P. P. K., Liu, W., Chen, D., Yeung, D. S., Zhang, F., Wang, X., & Hsu, C. C. (2018). Face liveness detection using a flash against 2D spoofing attack. IEEE Transactions on Information Forensics and Security, 13(2), 521-534. https://doi.org/10.1109/TIFS.2017.2758748

European Committee for Standardization. (2024). Biometric data injection attack detection (CEN/TS 18099:2024). https://www.evs.ee/en/cen-ts-18099-2024

Firc, A., Drtil, D., Perešíni, M., Šalko, M., Staněk, V., & Malinka, K. (2026). Not all challenges are equal: Evaluating interactive liveness detection against real-time face swaps [Preprint]. arXiv. https://arxiv.org/abs/2610.16417

Gerstner, C. R., & Farid, H. (2022). Detecting real-time deep-fake videos using active illumination. In Proceedings of the IEEE/CVF Conference on Computer Vision and Pattern Recognition Workshops (pp. 53-60). https://doi.org/10.1109/CVPRW56347.2022.00015

Guo, H., Wang, X., & Lyu, S. (2023). Detection of real-time deepfakes in video conferencing with active probing and corneal reflection. In 2023 IEEE International Conference on Acoustics, Speech and Signal Processing (ICASSP). IEEE. https://doi.org/10.1109/ICASSP49357.2023.10094720

iProov. (n.d.). Patented concepts. https://www.iproov.com/iproov-system/technology/patented-concepts

Jia, S., Guo, G., Xu, Z., & Wang, Q. (2020). Face presentation attack detection in mobile scenarios: A comprehensive evaluation. *Image and Vision Computing*, 93, 103826. https://doi.org/10.1016/j.imavis.2019.11.004

Khan, A. A., Laghari, A. A., Inam, S. A., Ullah, S., Shahzad, M., & Syed, D. (2025). A survey on multimedia-enabled deepfake detection: State-of-the-art tools and techniques, emerging trends, current challenges & limitations, and future directions. *Discover Computing*, 28, 48. https://doi.org/10.1007/s10791-025-09550-0

Kurmankhojayev, D., Shadrikov, A., Gordin, D., Shkorin, M., Gabdullin, D., Kambetbayeva, A., & Kuatov, K. (2025). Virtual camera detection: Catching video injection attacks in remote biometric systems [Preprint]. arXiv. https://arxiv.org/abs/2512.10653

Lai, Z., Zhang, Y., Li, D., & Ni, J. (2025). Leveraging high-frequency diversified augmentation for general deepfake detection. *Journal of Information Security and Applications*, 89, 103994. https://doi.org/10.1016/j.jisa.2025.103994

Lai, Z., Zhang, Y., Li, D., Li, J., & Ni, J. (2026). IIR-Net: Illumination inconsistency representation with relighting-guided enhancement for generalizable deepfake detection. IEEE Transactions on Multimedia. Advance online publication. https://doi.org/10.1109/TMM.2026.3736517

Liu, H., Li, Z., Xie, Y., Jiang, R., Wang, Y., Guo, X., & Chen, Y. (2020). LiveScreen: Video chat liveness detection leveraging skin reflection. In IEEE INFOCOM 2020: IEEE Conference on Computer Communications (pp. 1083-1092). IEEE. https://doi.org/10.1109/INFOCOM41043.2020.9155400

Michael, P. F., Hao, Z., Belongie, S., & Davis, A. (2025). Noise-coded illumination for forensic and photometric video analysis. ACM Transactions on Graphics, 44(5), 165. https://doi.org/10.1145/3742892

Mittal, G., Hegde, C., & Memon, N. (2024). GOTCHA: Real-time video deepfake detection via challenge-response. In 2024 IEEE 9th European Symposium on Security and Privacy (EuroS&P) (pp. 1-20). IEEE. https://doi.org/10.1109/EuroSP60621.2024.00009

Ramachandra, R., & Busch, C. (2017). Presentation attack detection methods for face recognition systems: A comprehensive survey. *ACM Computing Surveys*, 50(1), 8, 1–37. https://doi.org/10.1145/3038924

Shang, J., & Wu, J. (2020). Protecting real-time video chat against fake facial videos generated by face reenactment. In 2020 IEEE 40th International Conference on Distributed Computing Systems (ICDCS) (pp. 689-699). IEEE. https://doi.org/10.1109/ICDCS47774.2020.00082

Shu, X., Li, X., Zuo, X., Xu, D., & Shi, J. (2023). Face spoofing detection based on multi-scale color inversion dual-stream convolutional neural network. *Expert Systems with Applications*, 224, 119988. https://doi.org/10.1016/j.eswa.2023.119988

Smith, D. F., Wiliem, A., & Lovell, B. C. (2015). Face recognition on consumer devices: Reflections on replay attacks. IEEE Transactions on Information Forensics and Security, 10(4), 736-745. https://doi.org/10.1109/TIFS.2015.2398819

Tang, D., Zhou, Z., Zhang, Y., & Zhang, K. (2018). Face flashing: A secure liveness detection protocol based on light reflections. In Proceedings of the 25th Annual Network and Distributed System Security Symposium (NDSS 2018). Internet Society. https://doi.org/10.14722/ndss.2018.23176

Vijaykumar, R., Purnapatra, S., Plesh, R., Imtiaz, M. H., Hou, D., & Schuckers, S. (2024). Deepfake attacks on biometric recognition: Evaluation of resistance to injection attacks. In 2024 IEEE 15th Annual Ubiquitous Computing, Electronics & Mobile Communication Conference (UEMCON) (pp. 404-410). IEEE. https://doi.org/10.1109/UEMCON62879.2024.10754732

Yu, Z., Qin, Y., Li, X., Zhao, C., Lei, Z., & Zhao, G. (2023). Deep learning for face anti-spoofing: A survey. IEEE Transactions on Pattern Analysis and Machine Intelligence, 45(5), 5609–5631. https://doi.org/10.1109/TPAMI.2022.3215850

# Appendix

This appendix presents the calibration experiments used to configure the color wave method. The calibration dataset comprised 90 bona fide sessions, divided into two subsets for stimulus-color palette selection (A1) and stimulus-duration selection (A2). All calibration sessions were also used for response-color predictor training (A3).

## A1 Stimulus color palette selection

We compared five candidate stimulus palettes to balance color-prediction reliability with challenge diversity. The candidate colors comprised the RGB primaries, the additive secondary colors, and four intermediate hues (Table 4).

**Table 4.** Test stimulus colors.

| # | Color | Code | RGB |
|---|---|---|---|
| 1 | Red | R | (255, 0, 0) |
| 2 | Green | G | (0, 255, 0) |
| 3 | Blue | B | (0, 0, 255) |
| 4 | Yellow | Y | (255, 255, 0) |
| 5 | Cyan | C | (0, 255, 255) |
| 6 | Magenta | M | (255, 0, 255) |
| 7 | Orange | O | (255, 128, 0) |
| 8 | Yellow-green | YG | (128, 255, 0) |
| 9 | Azure | A | (0, 128, 255) |
| 10 | Violet | V | (128, 0, 255) |

The RGB three-color palette contained the primaries, the baseline six-color palette added the secondary colors, and the two eight-color palettes added either the warm/green or blue/violet hues. The full palette contained all ten colors.

Six subjects completed 24 bona fide sessions, with at least three sessions per subject. Each session presented all ten candidate colors in a fixed order, with each stimulus lasting 1,000 ms. This duration allowed the facial response to develop and provided sufficient stable frames for color-prediction evaluation.

Fig. 21 shows the predicted facial-response color scores across the fixed ten-color calibration sequence. The scores generally follow the displayed sequence, although neighboring hues exhibit greater prediction ambiguity.

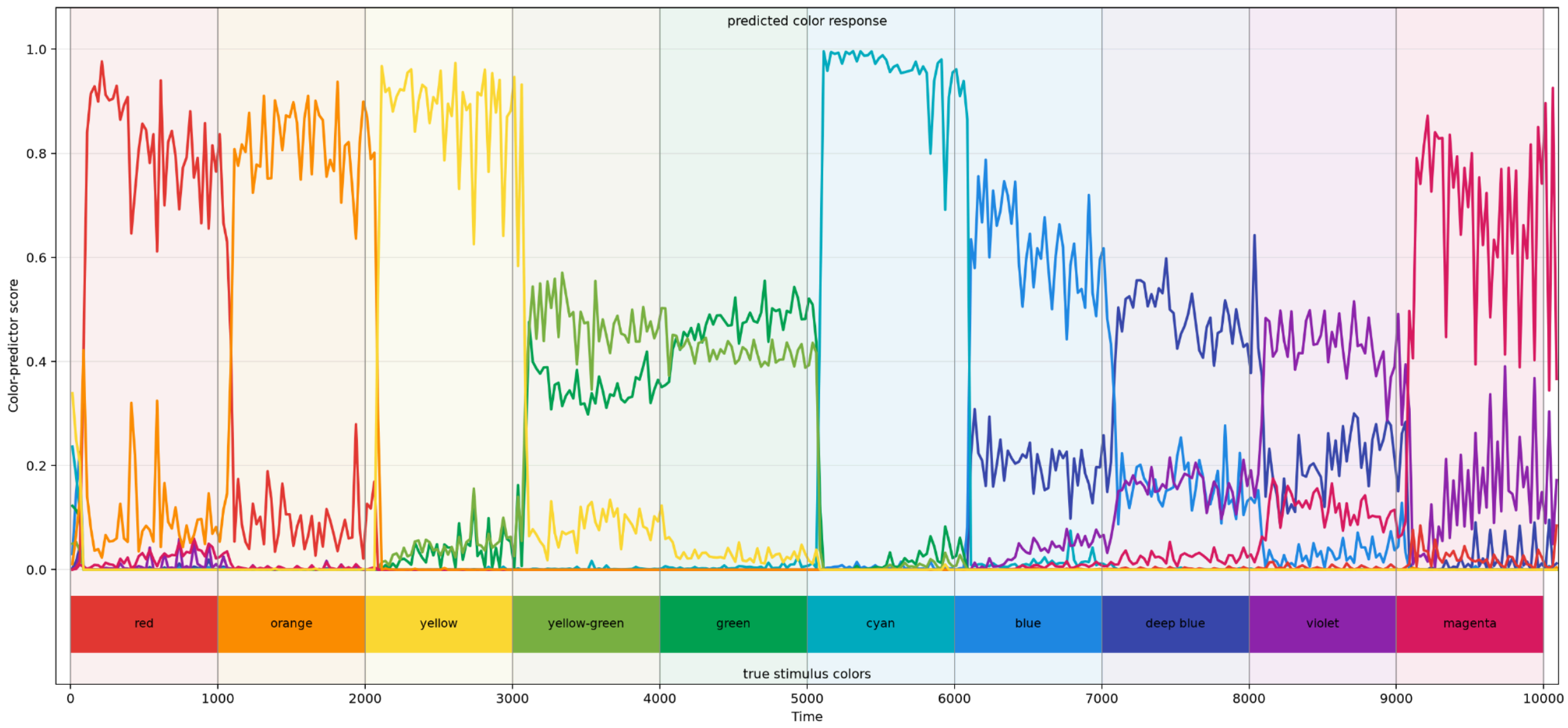

**Figure 21.** Predicted facial-response color scores across the fixed-order ten-color bona fide calibration sequence.

For each palette, a separate predictor was trained and evaluated using six-fold leave-one-subject-out cross-validation, with all sessions from one subject held out in each fold. The architecture and hyperparameters were fixed in advance, and attack data were not used for palette selection.

Table 5 reports frame-pair color-prediction performance rather than end-to-end attack-detection results. AUC is calculated from multiclass probability outputs, while accuracy is based on the highest-probability class.

**Table 5.** Color-prediction performance for alternative stimulus palettes.

| Palette | Classes | AUC | Accuracy |
|---|---|---|---|
| RGB three | 3 | 0.993 | 0.934 |
| Baseline six | 6 | **0.960** | **0.782** |
| Warm/green | 8 | 0.948 | 0.699 |
| Blue/magenta | 8 | 0.937 | 0.672 |
| Full ten | 10 | 0.941 | 0.627 |

The six-color RGBYCM palette was selected for the main experiments because it provided a practical compromise between prediction reliability and challenge diversity.

## A2 Stimulus duration

### A2.1 Minimum and maximum stimulus durations

We required at least five valid frames per stimulus to avoid estimating response durations from excessively sparse observations. This is a practical heuristic, not a physiological threshold.

For a stimulus with onset $t_i^{start}$, end time $t_i^{end}$, and fixed empirical reference response delay $L_{ref}$, the response window is defined by Eq. (5). In the present experiments, $L_{ref} = 100$ ms. At an effective frame rate $FPS$, the theoretical duration needed for five frames is:

$$D_{min} = \frac{5}{FPS} \times 1000 \text{ ms} \tag{12}$$

At 24 FPS, five frames correspond to approximately 208 ms. We selected 350 ms as the shortest candidate to provide additional tolerance for lower frame rates, dropped frames, and timestamp uncertainty. It provides approximately 5.6 frames at 16 FPS, 8.4 frames at 24 FPS, and 10.5 frames at 30 FPS. A stimulus is usable only when its fixed-reference response window contains at least five valid frames.

Candidate durations up to 950 ms were evaluated to examine the trade-off between duration distinguishability and session length. The final operational range was selected based on the comparison presented in A2.2.

Because the stimuli fill the screen, photosensitivity remains a deployment concern. Consecutive stimuli change color approximately 1.1–2.9 times per second. Although this nominal transition rate is below three changes per second, it does not establish compliance with WCAG 2.2 SC 2.3.1[3] or ISO 9241-391:2016[4]. Compliance also depends on the characteristics of visual transitions, including luminance and saturated-red changes, which were not formally evaluated. Deployment therefore requires a safety review, user warnings, and an opt-out procedure.

## A2.2 Candidate stimulus duration set selection

We compared three spacings between stimulus durations: 100, 200, and 300 ms. The duration-calibration subset comprised 66 bona fide sessions, with at least nine sessions collected for each of the seven candidate durations. Each session presented six colored stimuli of the same nominal duration. The candidate sets were:

- *100 ms spacing:* 350, 450, 550, 650, 750, 850, and 950 ms
- *200 ms spacing:* 350, 550, 750, and 950 ms
- *300 ms spacing:* 350, 650, and 950 ms

For each stimulus, we counted the frames whose predicted color matched the displayed color within the response window and converted the count to milliseconds using the session frame rate. Fig. 22 summarizes the corresponding measurements across candidate stimulus durations.

[3] Web Content Accessibility Guidelines (WCAG) 2.2, Success Criterion 2.3.1, "Three Flashes or Below Threshold." W3C Recommendation, 2024. https://www.w3.org/TR/WCAG22/#three-flashes-or-below-threshold

[4] ISO 9241-391:2016, Ergonomics of human-system interaction - Part 391: Requirements, analysis and compliance test methods for the reduction of photosensitive seizures. https://www.iso.org/standard/56350.html

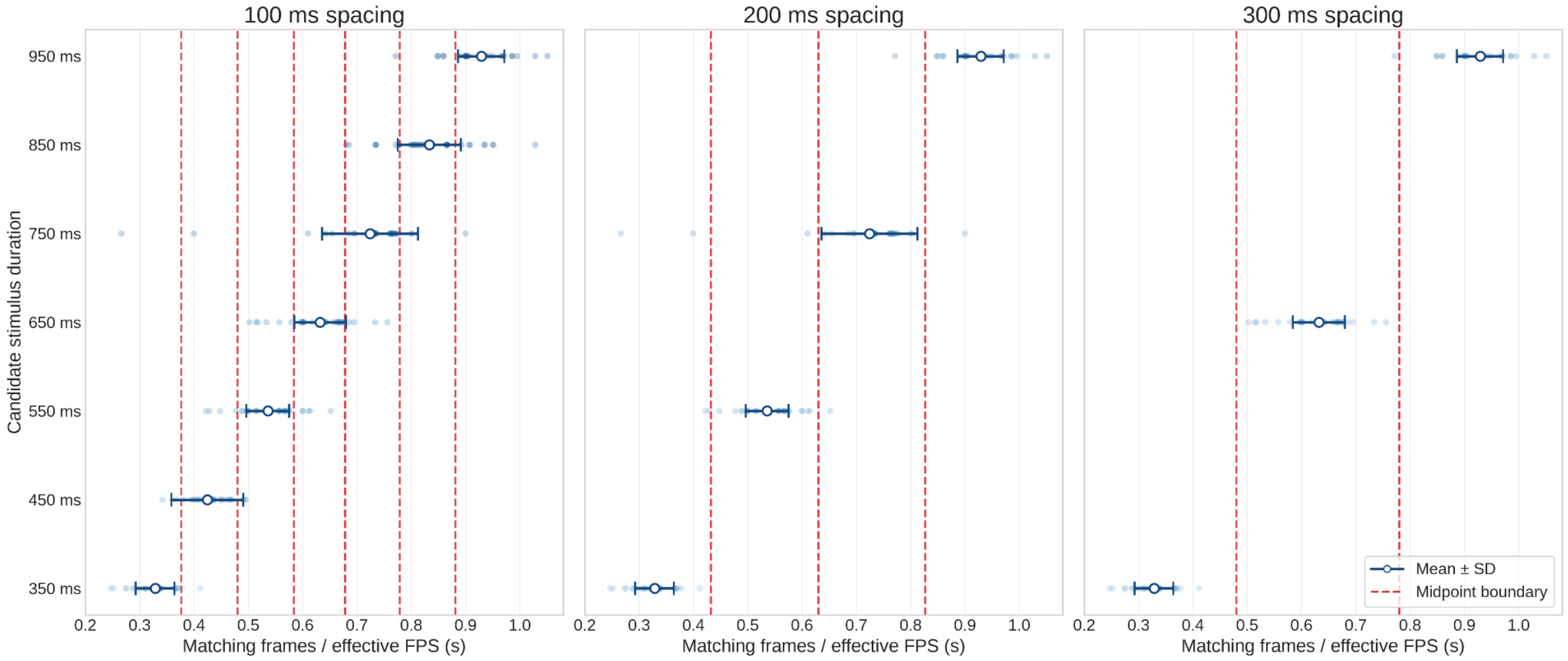

**Figure 22.** Estimated matching-color duration by nominal stimulus duration.

Separation was measured as the proportion of stimuli whose estimated response durations were correctly mapped to their nominal durations. The 100 ms spacing yielded 89–95% separation, indicating greater overlap between neighboring durations. The 300 ms spacing achieved 98–100% separation but offered fewer duration choices. The 200 ms spacing provided a practical compromise, achieving 97–98% separation while retaining more duration choices.

Although 950 ms was included in the candidate comparison, it was excluded from the operational configuration to limit session duration and user burden. The main experiments therefore use the operational stimulus-duration set $\mathcal{D} = 350, 550, 750$ ms, with each duration appearing at least once in the six-stimulus challenge.

# A3 Response-color predictor

This section describes the training and evaluation of the response-color predictor used to estimate the color of the screen-induced facial response.

Each prediction uses two facial frames: a neutral reference frame and a frame captured during colored illumination. The model computes the pixel-level log-gain change between the frames and combines it with the neutral face image to form a six-channel input: three channels representing the change and three representing the reference image. This representation helps account for differences in baseline facial appearance and illumination. A CNN learns spatial patterns in the resulting facial response and assigns probabilities to the six candidate colors. The predictor output and response-frame selection are defined in Section 2.2.5, Eqs. (3) and (4).

The final predictor, cnn-aug, uses 48 × 48-pixel inputs and four convolutional stages. The predictor was trained using eligible frame pairs from all 90 bona fide calibration sessions, with AdamW optimization for 150 epochs, a batch size of 64, and a peak learning rate of 0.003. Performance was evaluated using five-fold session-grouped cross-validation, ensuring that frames from the same session were not split between training and validation folds. Following cross-validation, the final model was trained on all eligible calibration frame pairs.

The predictor achieved 94.7% accuracy and a pooled out-of-fold macro-AUROC of 0.9972, with a mean fold-level AUROC of 0.9969 ± 0.0046. These results supported its selection for the subsequent injection-detection experiments.